\documentclass{article}

\PassOptionsToPackage{numbers}{natbib}

    \usepackage[preprint]{neurips_2023}

\usepackage[utf8]{inputenc} %
\usepackage[T1]{fontenc}    %
\usepackage[colorlinks]{hyperref}       %
\usepackage{url}            %
\usepackage{nicefrac}       %
\usepackage{microtype}      %
\usepackage{graphicx}
\usepackage{amsmath}
\usepackage{amssymb}
\usepackage{booktabs}
\usepackage{microtype}
\usepackage{xspace}
\usepackage{amsmath,amsfonts,amssymb,amsthm}
\usepackage[dvipsnames]{xcolor}
\usepackage{diagbox}
\usepackage{pifont}
\usepackage{svg}
\usepackage{amsmath}
\usepackage{enumitem}
\usepackage{physics}
\usepackage{mathtools}
\usepackage{algorithm}
\usepackage{algorithmic}
\usepackage{gensymb}
\usepackage{wrapfig}
\usepackage{subcaption}
\usepackage{multirow}
\usepackage{footmisc}
\usepackage{empheq}
\usepackage{cases}
\usepackage{adjustbox}
\usepackage{array}
\usepackage{tabu}
\usepackage{flushend}
\usepackage{multirow}
\usepackage{bm}
\usepackage{color, colortbl}
\usepackage[numbers]{natbib}
\usepackage{makecell}

\hypersetup{
  citecolor=[HTML]{6AA68B},
  linkcolor=[HTML]{ED7D31},
}

\title{
When Do We Not Need Larger Vision Models?

}

\author{%
  Baifeng Shi \\
  UC Berkeley \\
  \And
  Ziyang Wu \\
  UC Berkeley \\
  \And
  Maolin Mao \\
  UC Berkeley \\
  \AND
  Xin Wang \\
  Microsoft Research \\
  \And
  Trevor Darrell \\
  UC Berkeley \\
}

\begin{document}

\makeatletter
\DeclareRobustCommand\onedot{\futurelet\@let@token\@onedot}
\def\@onedot{\ifx\@let@token.\else.\null\fi\xspace}

\def\eg{\emph{e.g.}\xspace} \def\Eg{\emph{E.g}\onedot}
\def\ie{\emph{i.e.}\xspace} \def\Ie{\emph{I.e}\onedot}
\def\vs{\emph{vs.}\xspace}
\def\cf{\emph{c.f}\onedot} \def\Cf{\emph{C.f}\onedot}
\def\etc{\emph{etc}\onedot} \def\vs{\emph{vs}\onedot}
\def\wrt{w.r.t\onedot} \def\dof{d.o.f\onedot}
\def\etal{\emph{et al}\onedot}
\def\viz{\emph{viz}\onedot}
\makeatother

\newcommand{\cmark}{\ding{51}}%
\newcommand{\xmark}{\ding{55}}%

\def\rvx{{\mathbf{x}}}
\def\rvy{{\mathbf{y}}}
\def\vx{{\bm{x}}}
\def\vy{{\bm{y}}}
\def\vmu{{\bm{\mu}}}
\def\mSigma{{\bm{\Sigma}}}
\def\mI{{\bm{I}}}

\newcommand{\stwo}{{S$^2$}\xspace}
\newcommand{\sthree}{{S$^3$}\xspace} 
\newcommand{\stwobf}{{\textbf{S$\mathbf{^2}$}}\xspace}
\newcommand{\sthreebf}{{\textbf{S$\mathbf{^3}$}}\xspace} 
\newcommand{\stwowrapper}{{S$^2$-Wrapper}\xspace}
\newcommand{\sthreewrapper}{{S$^3$-Wrapper}\xspace}

\definecolor{Gray}{gray}{0.9}
\definecolor{darkgreen}{RGB}{5, 73, 7}
\definecolor{lightgreen}{HTML}{9CCBB8}
\definecolor{lightred}{HTML}{E3242B}
\definecolor{lightorange}{HTML}{ED7D31}

\newcommand\minisection[1]{\vspace{1.3mm}\noindent \textbf{#1}}

\newcommand{\todo}[1]{{\color{purple}{\bf TODO: #1}}}
\newcommand{\td}[1]{{\color{blue}{\bf TD: #1}}}
\newcommand{\bs}[1]{{\color{red}{\bf BS: #1}}}

\setcounter{topnumber}{3}

\maketitle

\begin{abstract}
  Scaling up the size of vision models has been the \emph{de facto} standard to obtain more powerful visual representations. In this work, we discuss the point beyond which larger vision models are \textit{not} necessary. First, we demonstrate the power of \textbf{S}caling on \textbf{S}cales (\stwobf), whereby a pre-trained and frozen smaller vision model (\eg, ViT-B or ViT-L), run over multiple image scales, can outperform larger models (\eg, ViT-H or ViT-G) on classification, segmentation, depth estimation, Multimodal LLM (MLLM) benchmarks, and robotic manipulation. Notably, \stwo achieves state-of-the-art performance in detailed understanding of MLLM on the V$^\ast$ benchmark, surpassing models such as GPT-4V. We examine the conditions under which \stwo is a preferred scaling approach compared to scaling on model size. While larger models have the advantage of better generalization on hard examples, we show that features of larger vision models can be well approximated by those of multi-scale smaller models. This suggests most, if not all, of the representations learned by current large pre-trained models can also be obtained from multi-scale smaller models. Our results show that a multi-scale smaller model has comparable learning capacity to a larger model, and pre-training smaller models with \stwo can match or even exceed the advantage of larger models. We release a Python package that can apply \stwo on any vision model with \textit{one line of code}: \url{https://github.com/bfshi/scaling_on_scales}. 
\end{abstract}

\section{Introduction}

Scaling up  model size has been one of the key drivers of recent progress in various domains of artificial intelligence, including language modeling~\cite{brown2020language,radford2019language,touvron2023llama}, image and video generation~\cite{yu2022scaling,ramesh2022hierarchical,kondratyuk2023videopoet,videoworldsimulators2024}, \etc. Similarly, for visual understanding, larger models have consistently shown improvements across a wide range of  downstream tasks given sufficient pre-training data~\cite{tan2019efficientnet,zhai2022scaling,cherti2023reproducible,oquab2023dinov2}. This trend has led to the pursuit of gigantic models with up to tens of billions of parameters as a default strategy for achieving more powerful visual representations and enhanced performance on downstream tasks~\cite{cherti2023reproducible,dehghani2023scaling,sun2023eva,el2024scalable}.

\begin{figure}[t]
  \includegraphics[width=1\linewidth]{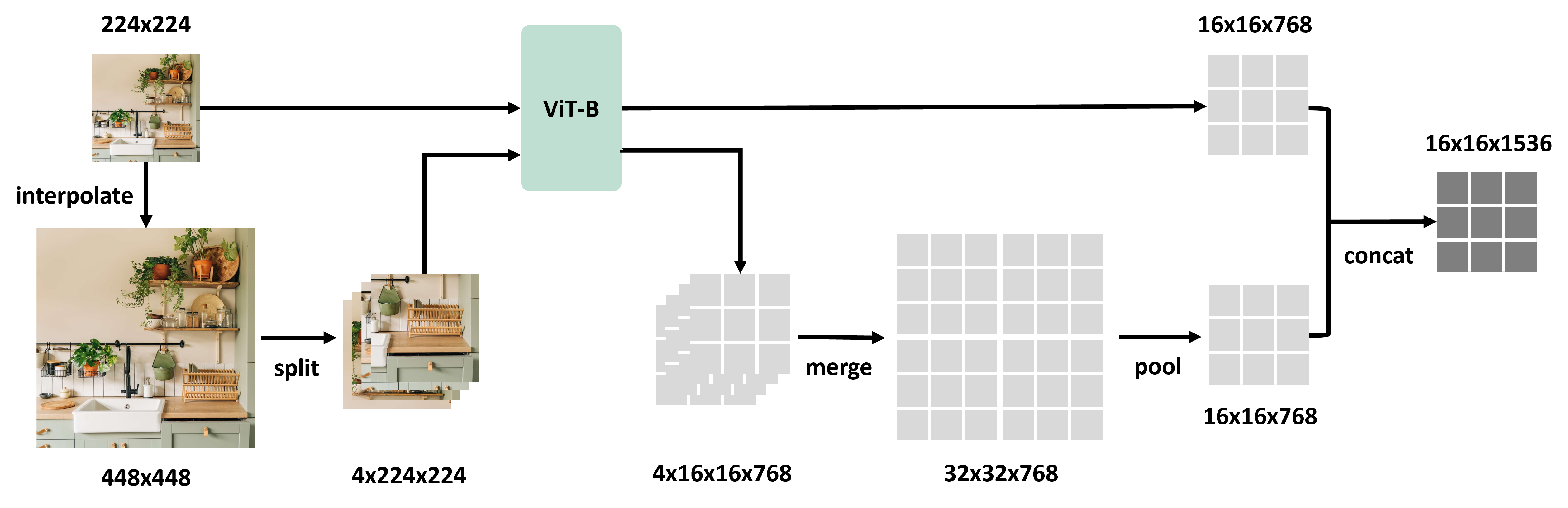}
  \vspace{-0.5em}
  \caption{\textbf{\stwowrapper is a simple mechanism that extends any pre-trained vision model to multiple image scales in a parameter-free manner.} Taking ViT-B as an example, \stwowrapper first interpolates the input image to different scales (\eg, $224^2$ and $448^2$) and splits each into several sub-images of the same size as the default input size ($448^2$ $\rightarrow$ $4 \times 224^2$). For each scale, all sub-images are fed into the same model and the outputs (\eg, $4 \times 16^2$) are merged into feature map of the whole image ($32^2$). Feature maps of different scales are average-pooled to the original spatial size ($16^2$) and concatenated together. The final multi-scale feature has the same spatial shape as single-scale feature while having higher channel dimension (\eg, 1536 \vs 768).}
  \label{fig:s2_wrapper}
\end{figure}

In this work, we revisit the question:  \textit{Is a larger model always necessary for better visual understanding?} 
Instead of scaling up model size, we consider scaling on the dimension of image scales---which we call \textbf{S}caling on \textbf{S}cales (\stwobf). With \stwo, a pre-trained and frozen smaller vision model (\eg, ViT-B or ViT-L) is run on multiple image scales to generate a multi-scale representation. We take a model pre-trained on single image scale (\eg, $224^2$), interpolate the image to multiple scales (\eg, $224^2$, $448^2$, $672^2$),  extract features on each scale by splitting larger images into sub-images of the regular size ($224^2$) and processing each separately before pooling them and concatenating with features from the original representation (Figure \ref{fig:s2_wrapper}). %

Surprisingly, from evaluations on visual representations of various pre-trained models (\eg, ViT~\cite{dosovitskiy2020image}, DINOv2~\cite{oquab2023dinov2}, OpenCLIP~\cite{cherti2023reproducible}, MVP~\cite{radosavovic2023real}), we show that smaller models with \stwo scaling consistently outperform larger models on classification, semantic segmentation, depth estimation, MLLM benchmarks, and robotic manipulation, with significantly fewer parameters ($0.28\times$ to $0.07\times$) and comparable GFLOPS. Remarkably, by scaling up image scale to $1008^2$, we achieve state-of-the-art performance in MLLM visual detail understanding on V$^\ast$ benchmark~\cite{vstar}, surpassing open-source and even commercial MLLMs like Gemini Pro~\cite{team2023gemini} and GPT-4V~\cite{achiam2023gpt}.

We further examine conditions under which \stwo is a preferred scaling approach compared to model size scaling. We find that while smaller models with \stwo achieve better downstream performance than larger models in many scenarios, larger models can still exhibit superior generalization on hard examples. This prompts an investigation into whether smaller models can achieve the same level of generalization capability as larger ones. Surprisingly, we find that the features of larger models can be well approximated by multi-scale smaller models through a single linear transform, which means smaller models should have at least a similar learning capacity of their larger counterparts. We hypothesize that their weaker generalization stems from being pre-trained with single image scale only. Through experiments of ImageNet-21k pre-training on ViT, we show that pre-training with \stwo scaling improves the generalizability of smaller models, enabling them to match or even exceed the advantages of larger models.

 \section{Related Work}

\minisection{Multi-scale representation} has been a common technique to recognize objects in a scale-invariant way since the era of feature engineering~\cite{dollar2014fast,dalal2005histograms,lowe2004distinctive} and is later introduced into convolutional neural networks~\cite{wang2020deep,lin2017feature,ronneberger2015u,tompson2015efficient} to extract features with both high-level semantics and low-level details. It has become a default test-time augmentation method for tasks such as detection and segmentation~\cite{mmseg2020,wu2019detectron2}, albeit at the cost of significantly slower inference speeds and typically limited image scales (up to $2\times$). Along with recent progress in vision transformers (ViT), variants of multi-scale ViTs~\cite{yang2021focal,fan2021multiscale,lee2022mpvit,chen2021crossvit} as well as hierarchical ViTs~\cite{liu2021swin,ryali2023hiera} have been proposed. However, these studies have not explored multi-scale representation as a general scaling approach as they usually design special architectures and are not applicable to common pre-trained vision models.

\minisection{Scaling Vision Models.} Training models with an increasing number of parameters has been the default approach to obtaining more powerful representations for visual pre-training~\cite{he2016deep,liu2022convnet,dosovitskiy2020image,oquab2023dinov2}. Previous research has studied how to optimally scale up vision models in terms of balancing model width, depth, and input resolution~\cite{tan2019efficientnet,tan2021efficientnetv2,bello2021revisiting,wightman2021resnet,dollar2021fast}, although they are usually limited to convolutional networks or even specific architectures such as ResNet~\cite{he2016deep}. Recent work also explores model size scaling of vision transformers in various settings~\cite{cherti2023reproducible,zhai2022scaling,dehghani2023scaling,riquelme2021scaling,bai2024sequential}. Others have incorporated high-resolution images into pre-training~\cite{oquab2023dinov2,fang2023eva,liu2022convnet,liu2021swin}, although the maximum resolution typically does not exceed $512^2$ due to unbearable demands of computational resources. Hu \etal \cite{hu2022exploring} study scaling on image scales through adjusting patch size for Masked Autoencoder (MAE)~\cite{he2022masked} where scaling is only applied on pre-training but not on downstream tasks.

\section{The Power of Scaling on Scales}
\label{sec:s2}

As an alternative to the conventional approach of scaling model size, we show the power of Scaling on Scales (\stwo), \ie, keeping the same size of a pre-trained model while running it on more and more image scales. From case studies on image classification, semantic segmentation, depth estimation, Multimodal LLMs, as well as robotic manipulation, we observe that \stwo scaling on a smaller vision model (\eg, ViT-B or ViT-L) often gives comparable or better performance than larger models (\eg, ViT-H or ViT-G), suggesting \stwo is a competitive scaling approach. In the following, we first introduce \stwowrapper, a mechanism that extends any pre-trained frozen vision model to multiple image scales without additional parameters (Section \ref{sec:s2_wrapper}). We then compare \stwo scaling and model size scaling in Section \ref{sec:s2_scaling_curve} - \ref{sec:sweetspot}.

\subsection{Scaling Pre-Trained Vision Models to Multiple Image Scales}
\label{sec:s2_wrapper}

We introduce \stwowrapper, a parameter-free mechanism to enable multi-scale feature extraction on any pre-trained vision model.  Regular vision models are normally pre-trained at a single image scale (\eg, $224^2$). \stwowrapper extends a pre-trained model to multiple image scales (\eg, $224^2$, $448^2$) by splitting different scales of images to the same size as seen in pre-training. Specifically, given the image at $224^2$ and $448^2$ scales, \stwowrapper first divides the $448^2$ image into four $224^2$ sub-images, which along with the original $224^2$ image are fed to the same pre-trained model. The features of four sub-images are merged back to the large feature map of the $448^2$ image, which is then average-pooled to the same size as the feature map of $224^2$ image. Output is the concatenation of feature maps across scales. The whole process is illustrated in Figure \ref{fig:s2_wrapper}. Note that instead of directly using the $448^2$ resolution image, we obtain the $448^2$ image by interpolating the $224^2$  image. This is to make sure no additional high-resolution information is introduced so we can make a fair comparison with model size scaling which never sees the high-resolution image. For practitioners, directly using the high-resolution image is recommended.

There are several key designs that make \stwowrapper efficient, effective, and easy to scale: (i) splitting the large image into small sub-images, instead of directly running on the whole large image, avoids quadratic computation complexity in self-attention and prevents performance degradation caused by position embedding interpolation~\cite{bolya2023window}, (ii) processing individual sub-images instead of using window attention allows using a pre-trained model that does not support window attention and avoids training additional parameters (\eg, relative position embedding) from scratch, (iii) interpolating the large feature map into the regular size makes sure the number of output tokens stays the same, preventing computational overhead in downstream applications such as MLLMs. Ablations of the designs can be found in Appendix~\ref{sec:appendix_ablation}. Note that we do not claim the novelty of extracting multi-scale features. Instead, we only choose the simplest algorithm design and study its scaling property.

\subsection{Scaling on Image Scales Can Beat Scaling on Model Size}
\label{sec:s2_scaling_curve}

\begin{figure}[t]
  \includegraphics[width=1\linewidth]{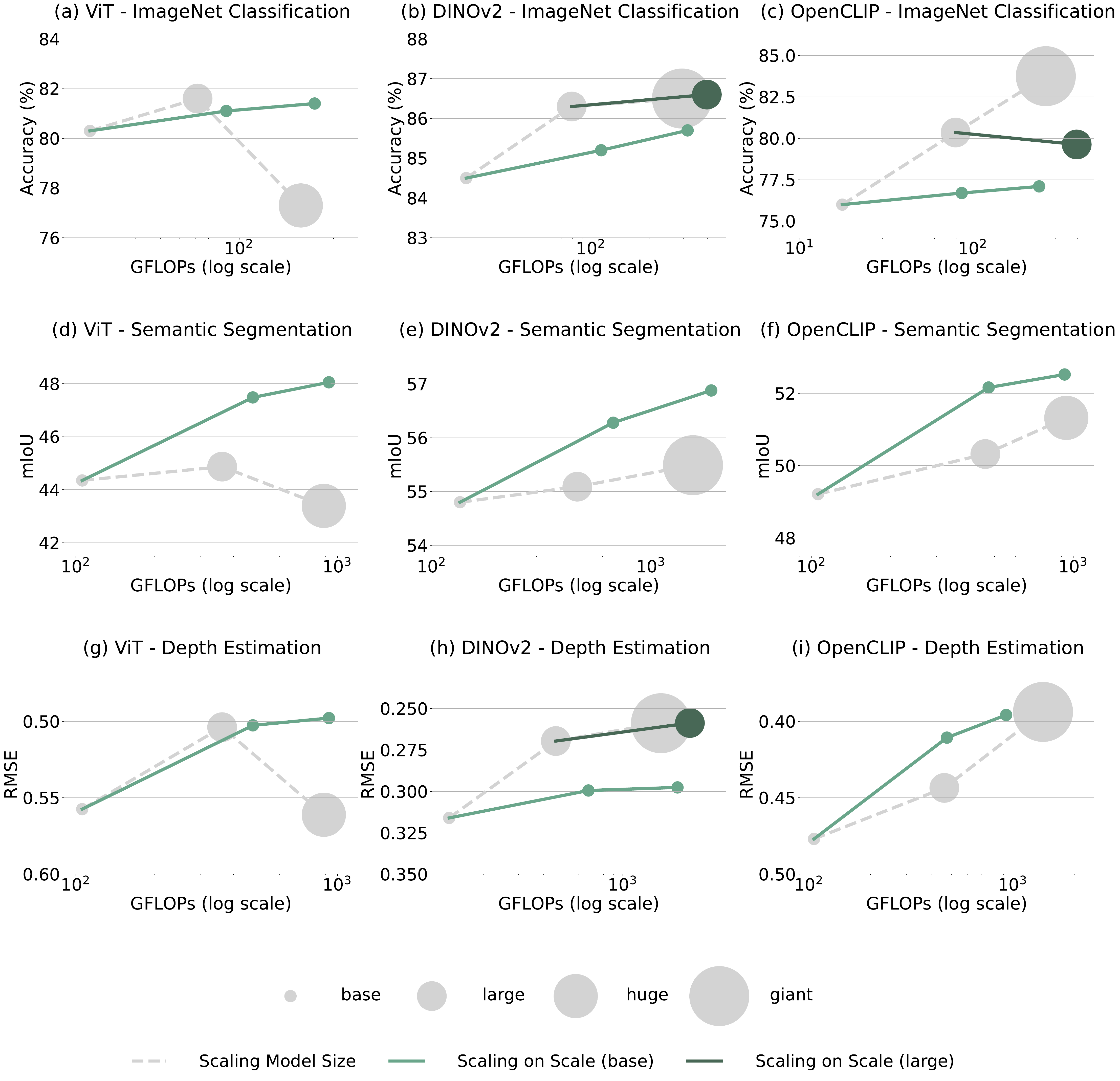}
  \vspace{-0.5em}
  \caption{\textbf{Comparison of \stwo scaling and model size scaling} on three models (ViT, DINOv2, and OpenCLIP) and three tasks (ImageNet classification, semantic segmentation, and depth estimation). For each model and each task, we test base, large, and huge/giant models for model size scaling (plotted in gray curve). For \stwo scaling (plotted in green curve), we test three sets of scales from single-scale (1x) to multi-scale (up to 3x), and we adjust each set of scale so that it matches the GFLOPs of the respective model size. Note that for specific models and tasks, we test \stwo scaling on both base and large models (plotted in light green and dark green curves separately). We can see that in (a), (d), (e), (f), (g), and (i), the base model with \stwo scaling already achieves comparable or better performances than larger models with similar GFLOPs and much smaller model size. For (b), (h), \stwo scaling from the large model is comparable with the giant model, again with similar GFLOPs and fewer parameters. The only failure case is (c), where \stwo scaling on either base or large models does not compete with model size scaling. }
  \label{fig:s2_scaling_curve}
\end{figure}

\stwowrapper enables \stwo scaling, \ie, keeping the same size of a pre-trained model while getting more and more powerful features by running on more and more image scales. Here we compare the scaling curve of \stwo to the regular approach of scaling up model size and show that \stwo scaling is a competitive, and in some cases, preferred scaling approach. To get a holistic analysis of two scaling approaches, we test their scaling curves on three representative tasks (image classification, semantic segmentation, and depth estimation) which correspond to the three dimensions of vision model capability~\cite{malik2016three}, as well as on MLLMs and robotic manipulation which reflect the comprehensive ability of visual understanding. 

\minisection{Case study: image classification, semantic segmentation, and depth estimation.} We use ImageNet~\cite{russakovsky2015imagenet}, ADE20k~\cite{zhou2017scene}, and NYUv2~\cite{silberman2012indoor} datasets for each task, respectively. We test on three families of pre-trained models (ViT~\cite{dosovitskiy2020image}, DINOv2~\cite{oquab2023dinov2}, and OpenCLIP~\cite{cherti2023reproducible}), spanning pre-training with different datasets (ImageNet-21k, LVD-142M, LAION-2B) and different pre-training objectives (supervised, unsupervised, and weakly-supervised). To see if the same observation holds for convolutional networks, we also test on ConvNeXt~\cite{liu2022convnet} (See Appendix~\ref{sec:appendix_convnext}). To fairly evaluate the representation learned from pre-training, we freeze the backbone and only train the task-specific head for all experiments. We use a single linear layer, Mask2former~\cite{cheng2022masked}, and VPD depth decoder~\cite{zhao2023unleashing} as decoder heads for three tasks, respectively. For model size scaling, we test the performance of base, large, and huge or giant size of each model on each task. For \stwo scaling, we test three sets of scales including (1x), (1x, 2x), (1x, 2x, 3x). For example, for ViT on ImageNet classification, we use three sets of scales: ($224^2$), ($224^2$, $448^2$), and ($224^2$, $448^2$, $672^2$), which have the comparable GFLOPs as ViT-B, ViT-L, and ViT-H, respectively. Note that the scales for specific models and tasks are adjusted to match the GFLOPS of respective model sizes. The detailed configurations for each experiment can be found in Appendix~\ref{sec:appendix_exp_setting}.

The scaling curves are shown in Figure \ref{fig:s2_scaling_curve}. We can see that in six out of nine cases ((a), (d), (e), (f), (g), (i)), \stwo scaling from base models gives a better scaling curve than model size scaling, outperforming large or giant models with similar GFLOPs and much fewer parameters. In two cases ((b) and (h)), \stwo scaling from base models has less competitive results than large models, but \stwo scaling from large models performs comparatively with giant models. The only failure case is (c) where both base and large models with \stwo scaling fail to compete with the giant model. Note that ViT-H is worse than ViT-L on all three tasks possibly due to the sub-optimal pre-training recipe~\cite{steiner2021train}. We observe that \stwo scaling has more advantages on dense prediction tasks such as segmentation and depth estimation, which matches the intuition that multi-scale features can offer better detailed understanding which is especially required by these tasks. For image classification, \stwo scaling is sometimes worse than model size scaling (\eg, multi-scale DINOv2-B \vs DINOv2-L). We hypothesize this is due to the weak generalizability of the base model feature because we observe that the multi-scale base model has a lower training loss than the large model despite the worse performance, which indicates overfitting. In Section \ref{sec:analysis_performance} we show that this can be fixed by pre-training with \stwo scaling as well.

\begin{figure}[t]
  \includegraphics[width=1\linewidth]{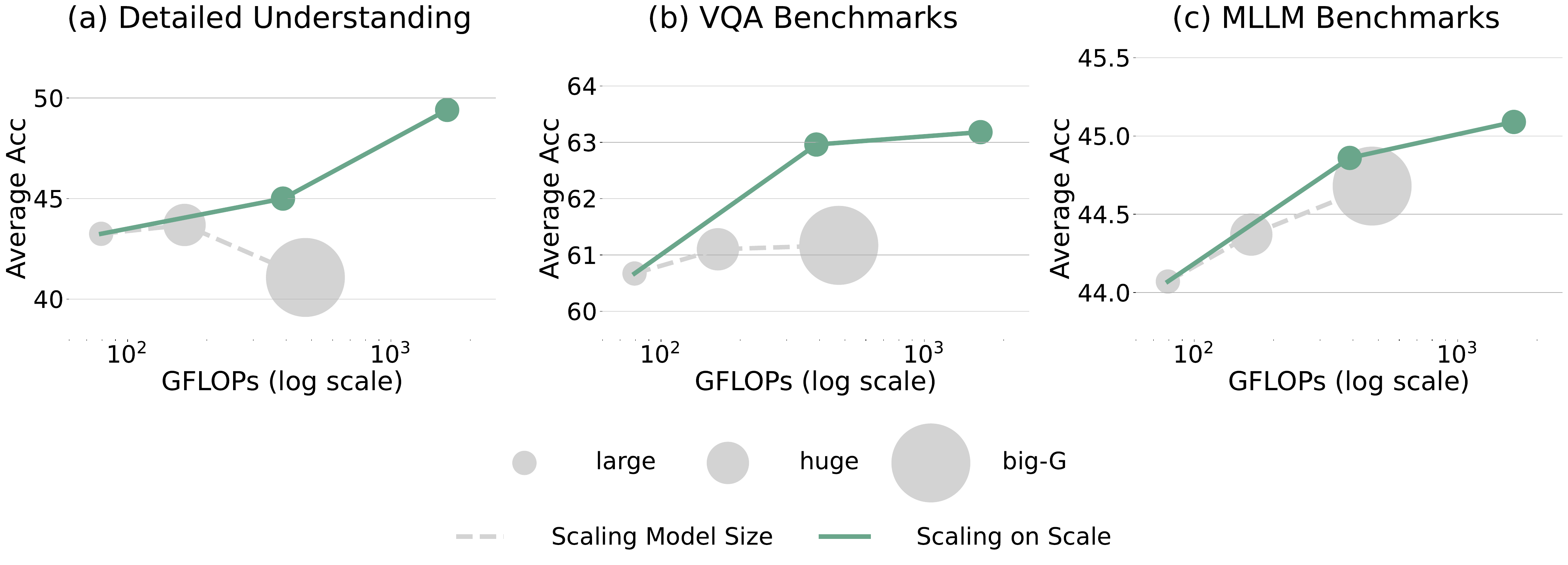}
  \vspace{-0.5em}
  \caption{\textbf{Comparison of \stwo scaling and model size scaling on MLLM.} \stwo scaling has comparable or better scaling curve than model size scaling on all three types of benchmarks. Using large image scales consistently gives better performance while using larger model can degrade model performance in certain cases. }
  \label{fig:vllm_scaling_curve}
\end{figure}

\minisection{Case study: Multimodal LLMs.} We compare \stwo scaling and model size scaling on MLLMs. We use a LLaVA~\cite{liu2023visual}-style model where LLM is a Vicuna-7B~\cite{chiang2023vicuna} and the vision backbone is OpenCLIP. We keep the same LLM and only change the vision backbone. For model size scaling, we test vision model sizes of large, huge, and big-G. For \stwo scaling, we keep the large-size model and test scales of ($224^2$), ($224^2$, $448^2$), and ($224^2$, $448^2$, $896^2$). For all experiments, we keep the vision backbone frozen and only train a projector layer between the vision feature and LLM input space as well as a LoRA~\cite{hu2021lora} on LLM. We follow the same training recipe as in LLaVA-1.5~\cite{liu2023improved}. We evaluate three types of benchmarks: (i) visual detail understanding (V$^\ast$~\cite{vstar}), (ii) VQA benchmarks (VQAv2~\cite{goyal2017making}, TextVQA~\cite{singh2019towards}, VizWiz~\cite{gurari2018vizwiz}), and (iii) MLLM benchmarks (MMMU~\cite{yue2023mmmu}, MathVista~\cite{lu2023mathvista}, MMBench~\cite{liu2023mmbench}, SEED-Bench~\cite{li2023seed}, MM-Vet~\cite{yu2023mm}).

\begin{figure}[ht]
    \centering
    \includegraphics[width=0.98\linewidth]{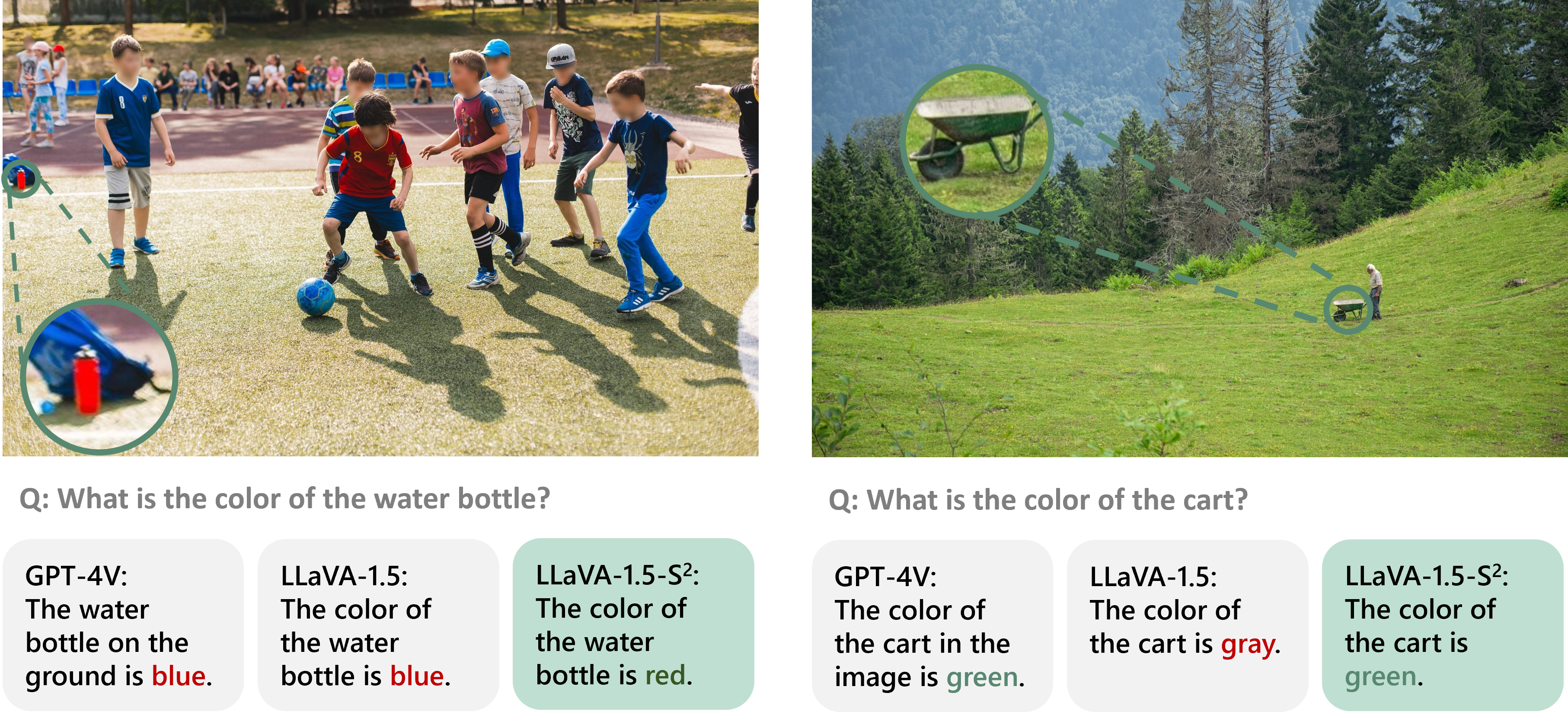} 
    \vspace{1em}
    \caption{\textbf{LLaVA-1.5 with \stwo scaling is able to recognize extremely fine-grained details in an image}, \eg, the color of a water bottle which lives in only 23$\times$64 pixels of a $2250\times1500$ image.}
    \label{fig:vllm_example}
\end{figure}

\begin{table}[t]
\caption{\textbf{Results on MLLM.} We evaluate three types of benchmarks: visual detail understanding (V$^\ast$~\cite{vstar}), VQA benchmarks (VQAv2~\cite{goyal2017making}, TextVQA~\cite{singh2019towards}, VizWiz~\cite{gurari2018vizwiz}), and MLLM benchmarks (MMMU~\cite{yue2023mmmu}, MathVista~\cite{lu2023mathvista}, MMBench~\cite{liu2023mmbench}, SEED-Bench~\cite{li2023seed}, MM-Vet~\cite{yu2023mm}). Notably, \stwo significantly improves the detailed understanding capability on V$^\ast$ benchmark, outperforming commercial models such as GPT-4V.}
    \label{tab:vllm}
    \vspace{1em}
    \centering
    \begin{scriptsize}
    \begin{tabular}{lp{0.03\textwidth}p{0.04\textwidth}p{0.035\textwidth}p{0.035\textwidth}|p{0.03\textwidth}p{0.03\textwidth}p{0.03\textwidth}|p{0.04\textwidth}p{0.03\textwidth}p{0.03\textwidth}p{0.03\textwidth}p{0.05\textwidth}}
        \toprule
        \multicolumn{3}{c}{}&\multicolumn{2}{c|}{\textbf{Visual Detail}}&\multicolumn{3}{c|}{\textbf{VQA Benchmarks}}&\multicolumn{5}{c}{\textbf{MLLM Benchmarks}}\\
        \multirow{2}{*}{Model} & \multirow{2}{*}{Res.} & \multirow{2}{*}{\#Token} & \multirow{2}{*}{V$^\ast_{\text{Att}}$} & \multirow{2}{*}{V$^\ast_{\text{Spa}}$} & \multirow{2}{*}{VQA$^{\text{v2}}$} & \multirow{2}{*}{VQA$^{\text{T}}$} &
        \multirow{2}{*}{
        Viz} &\multirow{2}{*}{MMMU} & \multirow{2}{*}{Math} & \multirow{2}{*}{MMB} & \multirow{2}{*}{SEED} & \multirow{2}{*}{MMVet}\\
        & & & & & & & & & & & \\
        \midrule
        \multicolumn{3}{l}{\emph{Commercial or proprietary models}}\\
        GPT-4V~\cite{achiam2023gpt} & - & - & \textbf{51.3} & 60.5 & 77.2 & 78.0 & - & \textbf{56.8} & \textbf{49.9} & 75.8 & 71.6 & \textbf{67.6} \\
        Gemini Pro~\cite{team2023gemini} & - & - & 40.9 & 59.2 & 71.2 & 74.6 & - & 47.9 & 45.2 & 73.6 & 70.7 & 64.3 \\
        Qwen-VL-Plus~\cite{qwen_vl_plus_2024} & - & - & - & - & - & \textbf{78.9} & - & 45.2 & 43.3 & - & - & - \\
        \midrule
         \multicolumn{3}{l}{\emph{Open-source models}}\\
        InstructBLIP-7B~\cite{dai2023instructblip} & 224 & - & 25.2 & 47.4 & - & 50.1 & 34.5 & - & - & 36.0 & - & 26.2 \\
        QwenVL-7B~\cite{bai2023qwen} & 448 & 1024 & - & - & 78.8 & 63.8 & 35.2 & - & - & 38.2 & - & - \\
        QwenVL-Chat-7B~\cite{bai2023qwen} & 448 & 1024 & - & - & 78.2 & 61.5 & 38.9 & - & - & 60.6 & - & - \\
        CogVLM-Chat~\cite{wang2023cogvlm} & 490 & 1225 & - & - & \textbf{82.3} & 70.4 & - & 41.1 & 34.5 & \textbf{77.6} & \textbf{72.5} & 51.1 \\
        LLaVA-1.5-7B~\cite{liu2023improved} & 336 & 576 & 43.5 & 56.6 & 78.5 & 58.2 & 50.0 & 36.2 & 25.2 & 64.3 & 65.7 & 30.5 \\
        \rowcolor{lightgreen!40} LLaVA-1.5-7B-\stwo & 1008 & 576 & \textbf{51.3} & 61.8 & 80.0 & 61.0 & 50.1 & 37.7 & 25.3 & 66.2 & 67.9 & 32.4 \\
        LLaVA-1.5-13B~\cite{liu2023improved} & 336 & 576 & 41.7 & 55.3 & 80.0 & 61.3 & 53.6 & 36.4 & 27.6 & 67.8 & 68.2 & 35.4 \\
        \rowcolor{lightgreen!40} LLaVA-1.5-13B-\stwo & 1008 & 576 & 50.4 & \textbf{63.2} & 80.9 & 63.1 & \textbf{56.0} & 37.4 & 27.8 & 67.9 & 68.9 & 36.4 \\
        \bottomrule
    \end{tabular}
    \end{scriptsize}
    \vspace{0.5em}
\end{table}

A comparison of the two scaling approaches is shown in Figure \ref{fig:vllm_scaling_curve}. We report the average accuracy on each type of benchmarks. We can see that on all three types of benchmarks, \stwo scaling on large-size models performs better than larger models, using similar GFLOPs and much smaller model sizes. Especially, scaling to $896^2$ improves the accuracy of detailed understanding by about $6\%$. On all benchmarks, larger image scales consistently improve performance while bigger models sometimes fail to improve or even hurt performance. These results suggest \stwo is a preferable scaling approach for vision understanding in MLLMs as well.

We also observe that LLaVA-1.5, when equipped with \stwo scaling, is already competitive or better than state-of-the-art open-source and even commercial MLLMs. Results are shown in Table \ref{tab:vllm}. Here we use OpenAI CLIP~\cite{radford2021learning} as the vision model for fair comparison. On visual detail understanding, LLaVA-1.5 with \stwo scaling outperforms all other open-source MLLMs as well as commercial models such as Gemini Pro and GPT-4V. This is credited to the highly fine-grained features we are able to extract by scaling image resolution to $1008^2$. A 
qualitative example is shown in Figure \ref{fig:vllm_example}. We can see that LLaVA-1.5 with \stwo is able to recognize an extremely small object that only takes $23 \times 64$ pixels in a $2250 \times 1500$ image and correctly answer the question about it. In the meantime, both GPT-4V and LLaVA-1.5 fail to give the correct answer. More qualitative examples are shown in Appendix~\ref{sec:appendix_qualitative}. On VQA and MLLM benchmarks, \stwo consistently improves the model performance as well, especially on benchmarks such as TextVQA which requires understanding of the fine details. Note that the improvement on certain MLLM benchmarks such as MathVista is not as significant as others, which is probably because these benchmarks require strong mathematical or reasoning capabilities which are not achievable by only improving vision but require stronger LLMs as well. In contrast to previous experiments, here we directly use the high-resolution image instead of interpolating from the low-resolution image in order to compare with the state of the arts.  Note that despite the large image scale, we keep the same number of image tokens as baseline LLaVA-1.5 since we interpolate the feature map of the large-scale images to the same size as that of the original image (see Section \ref{sec:s2_wrapper}). This makes sure the context length (and thus the computational cost) of LLM does not increase when using larger image scales, allowing us to use much higher resolution than the baselines. %

\newpage

\begin{wrapfigure}{r}{0.35\textwidth} %
  \centering
  \includegraphics[width=0.33\textwidth]{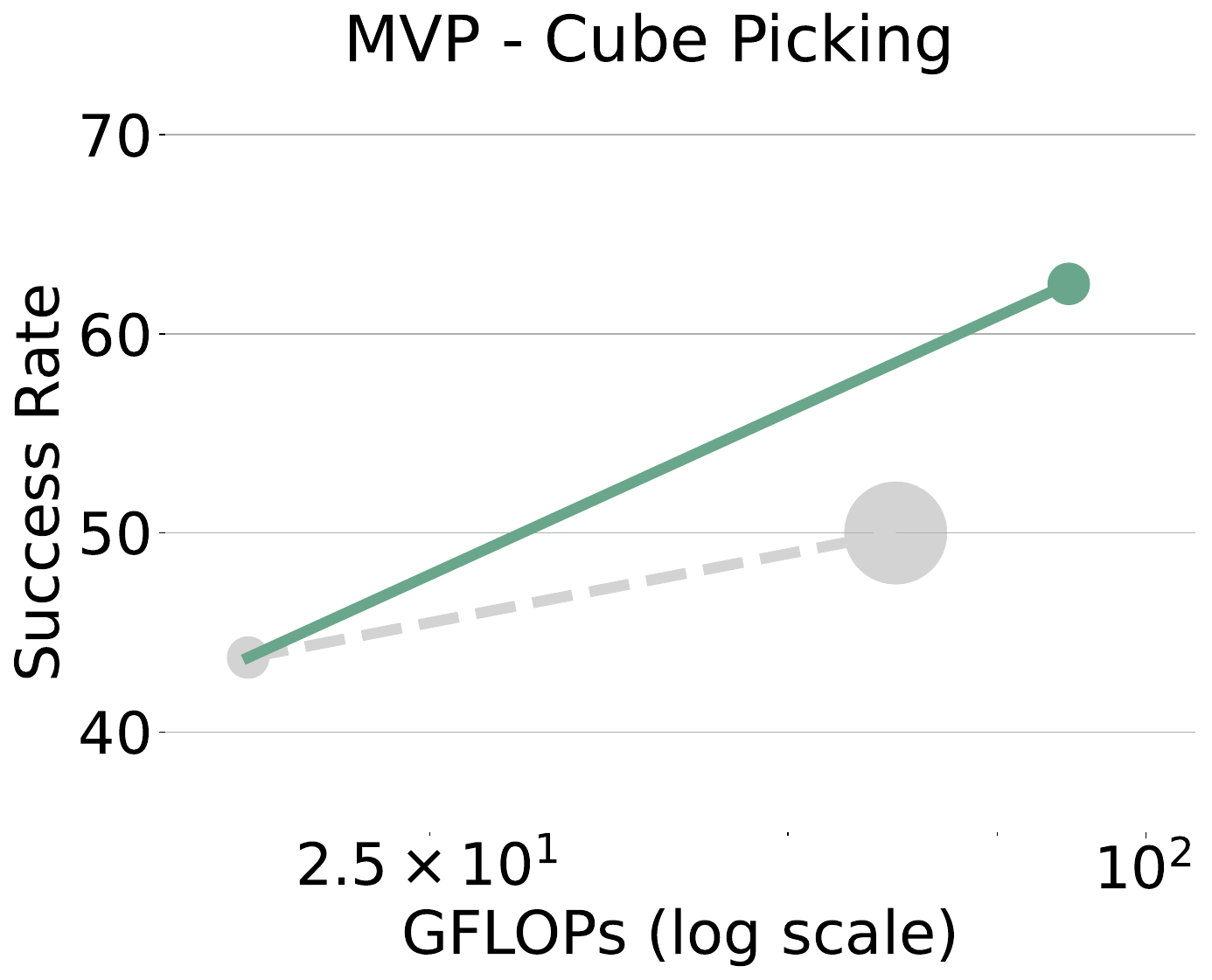} %
  \caption{\textbf{\stwobf \vs model size scaling on cube picking task.} \stwo scaling on base-size model improves the success rate by about $20\%$. } %
  \label{fig:robotics}
\end{wrapfigure}

\minisection{Case study: robotic manipulation.} We compare \stwo and model size scaling on a robotic manipulation task of cube picking. The task requires controlling a robot arm to pick up a cube on the table. We train a vision-based end-to-end policy on 120 demos using behavior cloning, and evaluate the success rate of picking on 16 randomly chosen cube positions, following the setting in \cite{radosavovic2023robot}. We use MVP~\cite{radosavovic2023real} as the pre-trained vision encoder to extract visual features which are fed to the policy. Please refer to Appendix~\ref{sec:appendix_exp_setting} for the detailed setting. To compare \stwo and model size scaling, we evaluate base and large models with single scale of ($224^2$), as well as a multi-scale base model with scales of ($224^2$, $448^2$). Results are shown in Figure \ref{fig:robotics}. Scaling from base to large model improves the success rate by about $6\%$, while scaling to larger image scales improves the success rate by about $20\%$. This demonstrates the advantage of \stwo over model size scaling on robotic manipulation tasks as well.

\subsection{The Sweet Spot Between Model Size Scaling and \stwo Scaling}
\label{sec:sweetspot}

\begin{figure}[t]
  \includegraphics[width=1\linewidth]{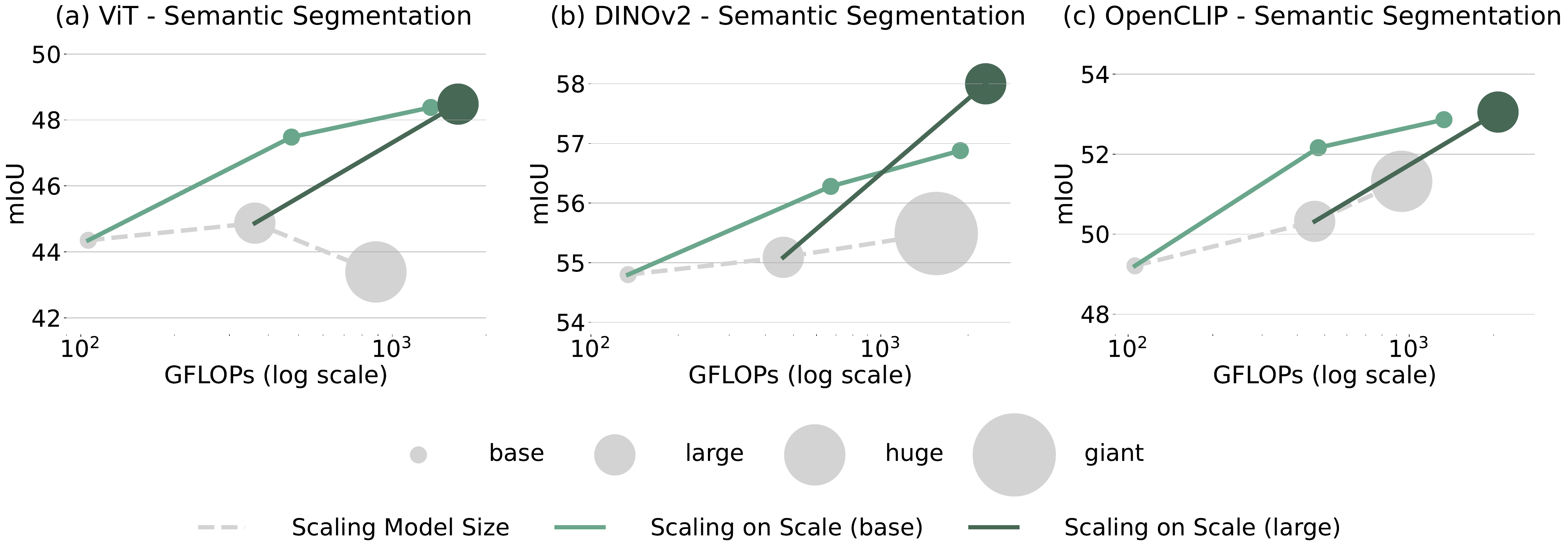}
  \vspace{-0.5em}
  \caption{\textbf{Which model size should we scale up image scales on?} The answer varies for different pre-trained models. For ViT and OpenCLIP, \stwo scaling from base or large model gives similar performances under computation budget beyond the huge-size model while the former performs better under similar GFLOPS as the large-size model. For DINOv2, \stwo scaling from the large size model has better performance than scaling from base size, especially under the same level of computation budget as the giant-size model.}
  \label{fig:sweetspot}
\end{figure}

While \stwo scaling outperforms model size scaling on a wide range of downstream tasks, a natural question arises: on which model size should we perform \stwo scaling? We show that it depends on different pre-trained models. For certain models, \stwo scaling from a large-size model gives an even better scaling curve when \stwo scaling from base model already beats larger models. As an example, we compare \stwo scaling from base and large models on semantic segmentation for ViT, DINOv2, and OpenCLIP. Results are shown in Figure \ref{fig:sweetspot}. We can see that for ViT and OpenCLIP, \stwo scaling from base models is better than from large models when the amount of computation is less than that of the huge-size models. These two curves eventually converge after going beyond the GFLOPs of the huge models. This means \stwo scaling from large models has no significant benefit than from base models. On the other hand, for DINOv2 we observe a clear advantage for \stwo scaling from the large model. When reaching the same level of GFLOPs as the giant-size model, \stwo scaling from the large model beats \stwo scaling from the base model by about 1 mIoU. These results indicate the optimal balancing between model size scaling and \stwo scaling varies for different models.

\section{The (Non)Necessity of Scaling Model Size}
\label{sec:analysis}

Results from Section \ref{sec:s2} suggest \stwo is a preferred scaling approach than model size scaling for various downstream scenarios. Nevertheless, larger vision models seem still necessary in certain cases (such as Figure \ref{fig:s2_scaling_curve}(c)) where \stwo scaling cannot compete with model size scaling. In the following, we first study the advantage of larger models and show they usually generalize better on rare or hard instances than multi-scale smaller models (Section \ref{sec:analysis_advantage}). Then, we explore if smaller models with \stwo scaling can achieve the same capability. We find that features of larger models can be well approximated by features of multi-scale smaller models, which means smaller models can learn what larger models learn to a large extent (Section \ref{sec:analysis_representation}). Based on this observation, we verify that multi-scale smaller models have similar capacity as larger models, and pre-training with \stwo scaling endows smaller models with similar or better generalization capability than larger models (Section \ref{sec:analysis_performance}).

\subsection{Larger Models Generalize Better on Hard Examples}
\label{sec:analysis_advantage}

\begin{figure}[t]
  \includegraphics[width=1\linewidth]{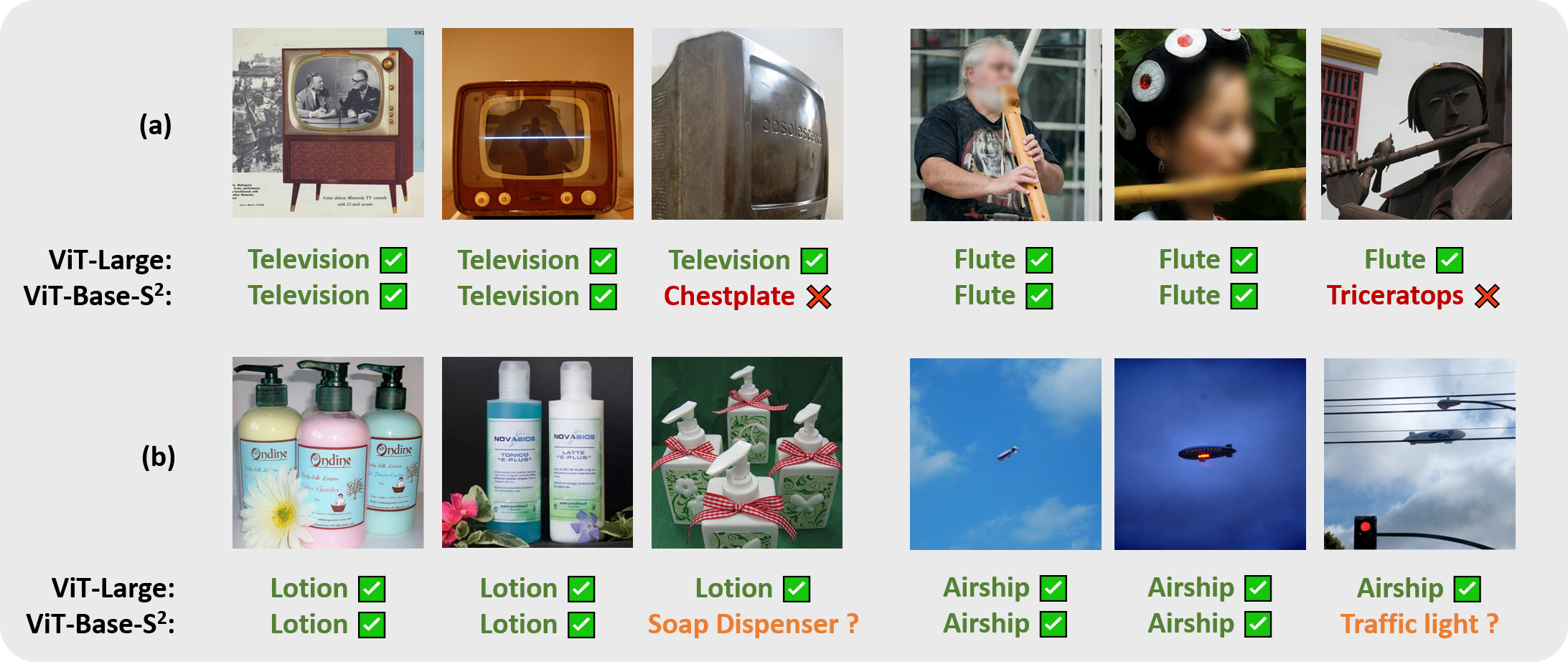}
  \vspace{-0.5em}
  \caption{\textbf{Types of samples that ViT-L improves the most but ViT-B-\stwo does not.} \textbf{(a) Rare cases}. These samples clearly belong to the class but are hard to classify due to the rare appearance (\eg, sculptures of television and flute). \textbf{(b) Ambiguous cases}. These samples have ambiguous labels. For example, the lotion could also be soap dispenser due to their high similarity, or the label could be either airship or traffic light when these two objects co-exist. }
  \label{fig:classification_compare}
\end{figure}

We use image classification as a testbed to understand the advantage of larger models. We conduct a qualitative analysis of what kinds of images are recognized better by a larger model but not by using larger image scales. Specifically, we find samples in ImageNet that a larger model (ViT-L) improves the most over a smaller model (ViT-B) but a multi-scale model (ViT-B-\stwo) fails to improve, as shown in Figure \ref{fig:classification_compare}. For each sample, we also find two easy samples (which two models both recognize correctly) from the same class as a comparison. We can see that there are mainly two types of images that larger models have advantages on. The first type is rare samples. For example, a television or a flute but in the form of a sculpture instead of regular ones (Figure \ref{fig:classification_compare}(a)). Larger models have larger capacity to learn to classify these rare examples during pre-training. The second type (Figure \ref{fig:classification_compare}(b)) is ambiguous examples, where the object can belong to either category (\eg, lotion and soap dispenser), or there are two categories co-existing in the same image and both labels should be correct (\eg, airship and traffic light). In this case, despite multiple correct labels, the large model is able to remember the label presented in the dataset during pre-training. While the second type is due to the flawed labeling process of ImageNet which makes it an unfair comparison and does not imply any disadvantage of multi-scale smaller models~\cite{beyer2020we,northcutt2021pervasive}, the first type indicates larger model can generalize better on rare or hard cases.

\subsection{Can Smaller Models Learn What Larger Models Learn?}
\label{sec:analysis_representation}

Is the advantage of larger models due to some unique representation they have learned that smaller models cannot learn? We design experiments to study how much of the representation of larger models is also learned by multi-scale smaller models. Surprisingly, our preliminary results suggest that \textit{most, if not all, of the representation of larger models is also learned by multi-scale smaller models.}

To quantify how much of the representation of a larger model (\eg, ViT-L) is also learned by a multi-scale smaller model (\eg, ViT-B-\stwo), we adopt a reconstruction-based evaluation, \ie, we train a linear transform to reconstruct the representation of a larger model from that of a multi-scale smaller model. Intuitively, low reconstruction loss means the representation of larger model can be equivalently learned by the multi-scale smaller model (through a linear transform) to a large extent. More formally, the reconstruction loss reflects the mutual information between two sets of representations. If we use MSE loss for reconstruction, the mutual information equals $I = - \log (l / l_0)$, where $l$ is the reconstruction loss and $l_0$ is the loss of vanilla reconstruction where the large model representation is reconstructed by a dummy vector (See Appendix~\ref{sec:appendix_derivation}). This quantifies how much information in the larger model representation is also contained in the multi-scale smaller model. We use a linear transform for reconstruction to (i) account for operations that keep the representation equivalence (e.g., channel permutation), (ii) measure the information that is useful for downstream tasks considering the task decoders are usually light-weight modules such as a single linear layer~\cite{xu2020theory}.

Moreover, in practice we find the reconstruction loss is usually nowhere near zero. We hypothesize this is because part of the feature is \textit{non-reconstructable} by nature, \ie, feature that is not relevant to the pre-training task and is learned due to randomness in weight initialization, optimization dynamics, \etc, thus cannot be reconstructed from another model's feature. To this end, we use an even larger (\eg, ViT-G) model to reconstruct the large model features as a comparison. Its reconstruction loss and corresponding mutual information are denoted by $l^\ast$ and $I^\ast = -\log (l^\ast / l_0)$. If we assume that, when pre-trained on the same task and the same dataset, any task-relevant feature learned by a smaller model can also be learned by a larger model, then all the useful features in a large-size model should be reconstructable by a huge or giant model as well. This means $I^\ast$, the amount of information reconstructed from a huge or giant model, should serve as an \textit{upper bound} of $I$. We empirically find this is indeed the case (see below). Therefore, we use the reconstruction ratio $I / I^\ast$ to measure how much representation in a larger model is also learned by a multi-scale smaller model.

We evaluate three classes of models: (i) ViT~\cite{dosovitskiy2020image} pre-trained on ImageNet-21k, (ii) OpenCLIP~\cite{cherti2023reproducible} pre-trained on LAION-2B, and (iii) MAE~\cite{he2022masked} pre-trained on ImageNet-1k. Reconstruction loss is averaged over all output tokens and is evaluated on ImageNet-1k. Results are shown in Table \ref{tab:feat_recon}. Compared to base models, we observe that multi-scale base models consistently have lower loss and reconstructs more information of the large model representation (\eg, 0.521 \vs 0.440 for ViT).  More interestingly, we find that the amount of information reconstructed from a multi-scale base model is usually close to that of a huge or giant model, although sometimes slightly lower but never exceeding by a large margin. For example, while OpenCLIP-Base reconstructs $92.7\%$ of the information, the multi-scale base model can reconstruct $99.9\%$. For other models, the reconstruction ratio of Base-\stwo model is usually close to $100\%$ while never exceeding by more than $0.5\%$. This implies (i) huge/giant models are indeed a valid upper bound of feature reconstruction, and (ii) most part of the feature of larger models is also learned by multi-scale smaller models. The only exception is when we reconstruct OpenCLIP-Huge feature, the reconstruction ratio is $88.9\%$. Although it's not near $100\%$, it is still significantly better than the base-size model which means at least a large part of the huge model feature is still multi-scale feature. These results imply smaller models with \stwo scaling should have at least a similar level of capacity to learn what larger models learn. On the other hand, we also notice that there exists a gap between train and test set, \ie, the reconstruction ratio on test set can be lower than train set (\eg $96.3\%$ \vs $99.9\%$ on OpenCLIP-L). We hypothesize this is because we only apply multi-scale after pre-training and the base model feature pre-trained on single image scale only has weaker generalizability.

\begin{table}[t]
    \caption{\textbf{Reconstructing representation of larger models from representation of regular or multi-scale smaller models.} We test three classes of models (ViT, OpenCLIP, and MAE), and for each class we test base, multi-scale base (Base-\stwo), and huge or giant model. We report results on both training and test set of ImageNet-1k, and for each we report the reconstruction loss, the amount of information reconstructed, and the percentage of information reconstructed compared to huge or giant model.}
    \label{tab:feat_recon}
    \vspace{1em}
    \centering
    \begin{small}
    \begin{tabular}{ccp{0.1\textwidth}<{\centering}cccccc}
        \toprule
        \multirow{2}{*}{Model Class} & \multirow{2}{*}{Target} & \multirow{2}{*}{Source} & \multicolumn{3}{c}{Train Set} & \multicolumn{3}{c}{Test Set} \\
         &  &  & Loss & Info & Ratio (\%) & Loss & Info & Ratio (\%) \\
        \midrule
        \multirow{3}{*}{ViT} & \multirow{3}{*}{Large} & Base & 0.1100 &  0.440 & 82.9\% & 0.0994 & 0.524 & 87.6\% \\
         &  & Base-\stwo & 0.1040 & 0.521 & \textbf{98.1\%} & 0.0942 & 0.601 & \textbf{100.5\%} \\
         &  & Huge & 0.1033 & 0.531 & 100\% & 0.0944 & 0.598 & 100\% \\
        
        \midrule
        \multirow{3}{*}{MAE} & \multirow{3}{*}{Large} & Base & 0.0013 & 7.460 & 97.3\% & 0.0010 & 7.840 & 96.0\% \\
         &  & Base-\stwo & 0.0011 & 7.694 & \textbf{100.3\%} & 0.0009 & 7.972 & \textbf{97.6\%} \\
         &  & Huge & 0.001 & 7.669 & 100\% & 0.0008 & 8.169 & 100\% \\
        \midrule
        \multirow{3}{*}{OpenCLIP} & \multirow{3}{*}{Large} & Base & 0.3693 & 1.495 & 92.7\% & 0.3413 & 1.723 & 90.7\% \\
         &  & Base-\stwo & 0.3408 & 1.611 & \textbf{99.9\%} & 0.3170 & 1.830 & \textbf{96.3\%} \\
         &  & Giant & 0.3402 & 1.613 & 100\% & 0.3022 & 1.900 & 100\% \\
        \midrule
        \multirow{3}{*}{OpenCLIP} & \multirow{3}{*}{Huge} & Base & 0.3926 & 1.407 & 83.2\% & 0.4231 & 1.413 & 80.8\% \\
         &  & Base-\stwo & 0.3670 & 1.504 & \textbf{88.9\%} & 0.3970 & 1.505 & \textbf{86.0\%} \\
         &  & Giant & 0.3221 & 1.692 & 100\% & 0.3354 & 1.749 & 100\% \\
        \bottomrule
    \end{tabular}
    \end{small}
\end{table}

\subsection{Pre-Training With \stwo Makes Smaller Models Better}
\label{sec:analysis_performance}

Given that most of the representation larger models have learned is also learned by multi-scale smaller models, we conjecture smaller models with \stwo scaling have at least similar capacity as larger models. Since larger capacity allows memorizing more rare and atypical instances during pre-training when given sufficient data and thus improves generalization error~\cite{feldman2020does,feldman2020neural,lukasik2023larger,cheng2022memorize,bartlett2020benign}, we further speculate smaller models can achieve similar or even better generalizability than larger models if pre-trained with \stwo scaling as well. We verify these in the following.

\begin{figure}[t]
    \centering
    \begin{minipage}{0.5\textwidth}
        \centering
        \captionof{table}{\textbf{Training loss on instance memorization and image classification.} A base model with \stwo scaling has similar memorization and classification losses, which implies it has at least the same level of model capacity as a large model.}
        \label{tab:capacity}
        \begin{small}
        \begin{tabular}{lccc}
            \toprule
                Model & Mem. Loss & \makecell{Cls. Loss \\ (DINOv2)} & \makecell{Cls. Loss \\ (OpenCLIP)} \\
            \midrule
                Base & 1.223 & 3.855 & 4.396 \\
                Large & \textbf{1.206} & 3.350 & \textbf{3.735} \\
                \rowcolor{lightgreen!40} Base-\stwo & \textbf{1.206} & \textbf{2.921} & 3.754 \\
            \bottomrule
        \end{tabular}
        \end{small}
    \end{minipage}%
    \hfill
    \begin{minipage}{0.43\textwidth}
        \centering
        \captionof{table}{\textbf{Pre-training with \stwobf.} Applying \stwo on a already pre-trained model has sub-optimal performance, while pre-training with \stwo makes smaller models better.}
        \label{tab:pretrain}
        \begin{small}
        \begin{tabular}{lccc}
            \toprule
                Model & \makecell{Pre-train \\ w/ \stwo}  & \makecell{Acc. \\ (ViT)} & \makecell{Acc. \\ (DINOv2)} \\
            \midrule
                Base &  & 80.3 & 77.6 \\
                Large &  & 81.6 & \textbf{81.9} \\
                \rowcolor{lightgreen!40} Base-\stwo & \xmark & 81.1 & 78.4 \\
                \rowcolor{lightorange!30} Base-\stwo & \cmark & \textbf{82.4} & 80.4 \\
            \bottomrule
        \end{tabular}
        \end{small}
    \end{minipage}
\end{figure}

\minisection{Multi-scale smaller models have similar capacity as larger models.} To measure the model capacity, we use two surrogate metrics: (i) memorization capability, and (ii) training loss on a specific task. For memorization capability, given a dataset (\eg, ImageNet), we regard each image as a separate category and train the model to classify individual images, which requires the model to memorize every single image. The classification loss reflects how well each instance is memorized and thus the model capacity~\cite{zhang2021understanding}. We adopt the training pipeline from \cite{wu2018unsupervised}. For training loss, we report classification loss on the training set of ImageNet-1k for DINOv2 and OpenCLIP. Lower loss means the model fits the training data better, which implies a larger model capacity.  Results are shown in Table \ref{tab:capacity}. For instance memorization, we can see that ViT-B with \stwo scaling ($224^2$ and $448^2$) has a similar loss as ViT-L. For ImageNet classification, ViT-B-\stwo has a similar training loss as ViT-L for OpenCLIP, and an even lower loss for DINOv2. These results suggest that multi-scale smaller models have at least comparable model capacity as larger models.

\minisection{Pre-training with \stwo makes smaller models better.} We evaluate ImageNet classification of a base model scaled with \stwo either during pre-training or after pre-training. We pre-train the model on ImageNet-21k, using either ViT image classification or DINOv2 as the pre-training objective. We compare models with or without \stwo during pre-training with single-scale base and large models. Results are shown in Table \ref{tab:pretrain}. We can see that when the base models are trained with single image scale and only scaled to multiple image scales after pre-training, they have sub-optimal performances compared to the large models, which aligns with our observation in Section \ref{sec:s2_scaling_curve}. However, when adding \stwo scaling into pre-training, the multi-scale base model is able to outperform the large model on ViT. For DINOv2, the base model pre-trained with \stwo achieves a performance that is significantly improved over the base model pre-trained without \stwo, and is more comparable to the large model. Although it still slightly falls behind the large model, pre-training a large model with \stwo potentially can give  a better scaling curve. These observations confirm our speculation that smaller models pre-trained with \stwo can match the advantage of larger models.

\section{Discussion}
\label{sec:discussion}

In this work, we ask the question \textit{is a larger model always necessary for better visual understanding?} We find that scaling on the dimension of image scales---which we call Scaling on Scales (\stwo)---instead of model size usually obtains better performance on a wide range of downstream tasks. We further show that smaller models with \stwo can learn most of what larger models learn, and pre-training smaller models with \stwo can match the advantage of larger models and even perform better. \stwo has a few implications for future work, including \textbf{(i) scale-selective processing}, \ie, not every scale at every position in an image contains equally useful features, and depending on image content and high-level task, it is much more efficient to select certain scales to process for each region, which resembles the bottom-up and top-down selection mechanism in human visual attention~\cite{li2014understanding,shi2023top,itti2001computational}, \textbf{(ii) parallel processing of single image}, \ie, in contrast with regular ViT where the whole image is processed together at once, the fact that each sub-image is processed independently in \stwo enables parallel processing of different sub-images for a single image, which is especially helpful for scenarios where latency on processing single large images is critical~\cite{zhang2021elf}.

\minisection{Acknowledgements.} We would like to thank Sheng Shen, Kumar Krishna Agrawal, Ritwik Gupta, Yossi Gandelsman, Chung Min Kim, Roei Herzig, Alexei Efros, Xudong Wang, and Ilija Radosavovic for their valuable discussions and suggestions on our project.

{\small
\bibliographystyle{plainnat}
\bibliography{egbib}
}

\newpage
\appendix
\section{Detailed Experimental Settings and Full Results}
\label{sec:appendix_exp_setting}

The details of the models and the corresponding results on image classification, semantic segmentation, and depth estimation are listed in Table \ref{tab:setting_classification}, \ref{tab:setting_segmentation}, and \ref{tab:setting_depth}, respectively. We use ImageNet-21k pre-trained checkpoints for ViT\footnote{\url{https://huggingface.co/google/vit-base-patch16-224-in21k}}\textsuperscript{,}\footnote{\url{https://huggingface.co/google/vit-large-patch16-224-in21k}}\textsuperscript{,}\footnote{\url{https://huggingface.co/google/vit-huge-patch14-224-in21k}}, LVD-142M pre-trained checkpoints for DINOv2\footnote{\url{https://dl.fbaipublicfiles.com/dinov2/dinov2_vitb14/dinov2_vitb14_pretrain.pth}}\textsuperscript{,}\footnote{\url{https://dl.fbaipublicfiles.com/dinov2/dinov2_vitl14/dinov2_vitl14_pretrain.pth}}\textsuperscript{,}\footnote{\url{https://dl.fbaipublicfiles.com/dinov2/dinov2_vitg14/dinov2_vitg14_pretrain.pth}}, and LAION-2B pre-trained checkpoints for OpenCLIP\footnote{\url{https://huggingface.co/laion/CLIP-ViT-B-16-laion2B-s34B-b88K}}\textsuperscript{,}\footnote{\url{https://huggingface.co/laion/CLIP-ViT-L-14-laion2B-s32B-b82K}}\textsuperscript{,}\footnote{\url{https://huggingface.co/laion/CLIP-ViT-g-14-laion2B-s34B-b88K}}.  For each model type (ViT~\cite{dosovitskiy2020image}, DINOv2~\cite{oquab2023dinov2}, OpenCLIP~\cite{cherti2023reproducible}), we choose the scales so that the models with \stwo have comparable number of FLOPs with corresponding larger models. For image classification, we train a linear classifier for $30$ epochs with learning rate of $0.0005$ and batch size of $512$. For semantic segmentation, we train a Mask2Former decoder~\cite{cheng2022masked} following the configurations here\footnote{\url{https://github.com/open-mmlab/mmsegmentation/blob/main/configs/mask2former/mask2former_r50_8xb2-160k_ade20k-512x512.py}}. For depth estimation, we train a VPD depth decoder~\cite{zhao2023unleashing} following the configurations here\footnote{\url{https://github.com/open-mmlab/mmsegmentation/blob/main/configs/vpd/vpd_sd_4xb8-25k_nyu-512x512.py}}.

\begin{table}[ht]
\caption{Configurations of models and corresponding results on ImageNet classification.}
\label{tab:setting_classification}
\vspace{1em}
\centering
\begin{small}
\begin{tabular}{llllll}
\toprule
 & Model Size & Scales & \#Params & \#FLOPs & Acc. \\
\midrule
\multirow{5}{*}{ViT} & Base & ($224^2$) & 86M & 17.6G & 80.3 \\
 & Base & ($224^2$, $448^2$) & 86M & 88.1G & 81.1 \\
 & Base & ($224^2$, $448^2$, $672^2$) & 86M & 246.0G & 81.4 \\
 & Large & ($224^2$) & 307M & 61.6G & 81.6 \\
 & Huge & ($224^2$) & 632M & 204.9G & 77.3 \\
\midrule
\multirow{6}{*}{DINOv2} & Base & ($224^2$) & 86M & 22.6G & 84.5 \\
 & Base & ($224^2$, $448^2$) & 86M & 112.8G & 85.2 \\
 & Base & ($224^2$, $448^2$, $672^2$) & 86M & 315.9G & 85.7 \\
 & Large & ($224^2$) & 303M & 79.4G & 86.3 \\
 & Large & ($224^2$, $448^2$) & 303M & 397.1G & 86.6 \\
 & Giant & ($224^2$) & 632M & 295.4G & 86.5 \\
 \midrule
\multirow{6}{*}{OpenCLIP} & Base & ($224^2$) & 86M & 17.6G & 76.0 \\
 & Base & ($224^2$, $448^2$) & 86M & 86.1G & 76.7 \\
 & Base & ($224^2$, $448^2$, $672^2$) & 86M & 241.0G & 77.1 \\
 & Large & ($224^2$) & 303M & 79.4G & 80.4 \\
 & Large & ($224^2$, $448^2$) & 303M & 397.1G & 79.6 \\
 & Giant & ($224^2$) & 1012M & 263.4G & 83.8 \\
 \bottomrule
\end{tabular}
\end{small}
\end{table}

Table \ref{tab:setting_mllm_vqa} and \ref{tab:setting_mllm_benchmark} show the model details and full results for V$^\ast$, VQA tasks, and MLLM benchmarks. We use OpenCLIP with large, huge, and big-G sizes, and also large-size model with $(224^2)$, $(224^2, 448^2)$, $(224^2, 448^2, 672^2)$ scales. We follow the training and testing configurations in LLaVA-1.5\footnote{\url{https://github.com/haotian-liu/LLaVA}}. For evaluations on certain MLLM benchmarks such as MMMU~\cite{yue2023mmmu}, since it is not supported in the LLaVA-1.5 repo, we use VLMEvalKit~\cite{2023opencompass} for evaluation\footnote{\url{https://github.com/open-compass/VLMEvalKit}}.

Table \ref{tab:setting_robotics} shows the model details and full results for the robotic manipulation task of cube picking. We use MVP~\cite{radosavovic2023real} as the vision backbone and use base and large size as well as base size with $(224^2, 448^2)$ scales. The vision backbone is frozen and extracts the visual feature for the visual observation at each time step. We train a transformer that takes in the visual features, proprioception and actions for the last 16 steps and outputs the actions for the next 16 steps. We train the model with behavior cloning on 120 self-collected demos. We test the model on 16 randomly selected cube positions and report the rate of successfully picking up the cube at these positions.

\begin{table}[ht]
\caption{Configurations of models and corresponding results on ADE20k semantic segmentation.}
\label{tab:setting_segmentation}
\vspace{1em}
\centering
\begin{small}
\begin{tabular}{llllll}
\toprule
 & Model Size & Scales & \#Params & \#FLOPs & mIoU \\
\midrule
\multirow{5}{*}{ViT} & Base & ($512^2$) & 86M & 105.7G & 44.4 \\
 & Base & ($256^2$, $512^2$, $1024^2$) & 86M & 474.7G & 47.8 \\
 & Base & ($256^2$, $512^2$, $1536^2$) & 86M & 926.7G & 48.0 \\
 & Large & ($512^2$) & 307M & 362.1G & 44.9 \\
 & Huge & ($512^2$) & 632M & 886.2G & 43.4 \\
\midrule
\multirow{5}{*}{DINOv2} & Base & ($518^2$) & 86M & 134.4G & 54.8 \\
 & Base & ($518^2$, $1036^2$) & 86M & 671.8G & 56.3 \\
 & Base & ($518^2$, $1036^2$, $1554^2$) & 86M & 1881G & 56.9 \\
 & Large & ($518^2$) & 303M & 460.9G & 55.1 \\
 & Giant & ($518^2$) & 632M & 1553G & 55.5 \\
 \midrule
\multirow{5}{*}{OpenCLIP} & Base & ($512^2$) & 86M & 105.7G & 49.2 \\
 & Base & ($256^2$, $512^2$, $1024^2$) & 86M & 474.7G & 52.2 \\
 & Base & ($256^2$, $512^2$, $1536^2$) & 86M & 926.7G & 52.6 \\
 & Large & ($518^2$) & 303M & 460.9G & 50.3 \\
 & Huge & ($518^2$) & 632M & 940.2G & 51.3 \\
 \bottomrule
\end{tabular}
\end{small}
\end{table}

\begin{table}[ht]
\caption{Configurations of models and corresponding results on NYUv2 depth estimation.}
\label{tab:setting_depth}
\vspace{1em}
\centering
\begin{small}
\begin{tabular}{llllll}
\toprule
 & Model Size & Scales & \#Params & \#FLOPs & RMSE \\
\midrule
\multirow{5}{*}{ViT} & Base & ($512^2$) & 86M & 105.7G & 0.5575 \\
 & Base & ($256^2$, $512^2$, $1024^2$) & 86M & 474.7G & 0.5127 \\
 & Base & ($256^2$, $512^2$, $1536^2$) & 86M & 926.7G & 0.5079 \\
 & Large & ($512^2$) & 307M & 362.1G & 0.5084 \\
 & Huge & ($512^2$) & 632M & 886.2G & 0.5611 \\
\midrule
\multirow{6}{*}{DINOv2} & Base & ($504^2$) & 86M & 134.4G & 0.3160 \\
 & Base & ($504^2$, $1008^2$) & 86M & 671.8G & 0.2995 \\
 & Base & ($504^2$, $1008^2$, $1512^2$) & 86M & 1881G & 0.2976 \\
 & Large & ($504^2$) & 303M & 460.9G & 0.2696 \\
 & Large & ($504^2$, $1008^2$) & 303M & 2170G & 0.2584 \\
 & Giant & ($504^2$) & 632M & 1553G & 0.2588 \\
 \midrule
\multirow{5}{*}{OpenCLIP} & Base & ($512^2$) & 86M & 105.7G & 0.4769 \\
 & Base & ($256^2$, $512^2$, $1024^2$) & 86M & 474.7G & 0.4107 \\
 & Base & ($256^2$, $512^2$, $1536^2$) & 86M & 926.7G & 0.3959 \\
 & Large & ($504^2$) & 303M & 460.9G & 0.4436 \\
 & Huge & ($504^2$) & 632M & 940.2G & 0.3939 \\
 \bottomrule
\end{tabular}
\end{small}
\end{table}

\begin{table}[ht]
\caption{Configurations of models and corresponding results on V$^\ast$ and VQA tasks.}
\label{tab:setting_mllm_vqa}
\vspace{1em}
\centering
\begin{scriptsize}
\begin{tabular}{llllllllll}
\toprule
 & Model Size & Scales & \#Params & \#FLOPs & V$^\ast_{\text{Att}}$ & V$^\ast_{\text{Spa}}$ & VQA$^{\text{v2}}$ & VQA$^{\text{T}}$ & Viz \\
\midrule
\multirow{5}{*}{OpenCLIP} & Large & ($224^2$) & 304M & 79.4G & 36.5 & 50.0 & 76.6 & 53.8 & 51.6 \\
 & Large & ($224^2$, $448^2$) & 304M & 389.1G & 40.0 & 50.0 & 77.8 & 55.9 & 55.2 \\
 & Large & ($224^2$, $448^2$, $672^2$) & 304M & 1634G & 35.7 & 63.2 & 77.9 & 56.5 & 55.3 \\
 & Huge & ($224^2$) & 632M & 164.6G & 37.4 & 50.0 & 76.0 & 54.0 & 53.3 \\
 & big-G & ($224^2$) & 1012M & 473.4G & 32.2 & 48.7 & 76.2 & 54.0 & 53.5 \\
 \bottomrule
\end{tabular}
\end{scriptsize}
\end{table}

\begin{table}[ht]
\caption{Configurations of models and corresponding results on MLLM  benchmarks.}
\label{tab:setting_mllm_benchmark}
\vspace{1em}
\centering
\begin{scriptsize}
\begin{tabular}{llllllllll}
\toprule
 & Model Size & Scales & \#Params & \#FLOPs & MMMU & Math & MMB & SEED & MMVet \\
\midrule
\multirow{5}{*}{OpenCLIP} & Large & ($224^2$) & 304M & 79.4G & 35.4 & 24.0 & 64.2 & 65.5 & 31.6 \\
 & Large & ($224^2$, $448^2$) & 304M & 389.1G & 37.6 & 24.2 & 64.5 & 66.0 & 33.0 \\
 & Large & ($224^2$, $448^2$, $672^2$) & 304M & 1634G & 37.8 & 24.5 & 64.0 & 66.3 & 32.8 \\
 & Huge & ($224^2$) & 632M & 164.6G & 36.1 & 25.2 & 64.2 & 65.6 & 30.7 \\
 & big-G & ($224^2$) & 1012M & 473.4G & 35.6 & 25.2 & 64.8 & 65.1 & 32.8 \\
 \bottomrule
\end{tabular}
\end{scriptsize}
\end{table}

\begin{table}[ht]
\caption{Configurations of models and corresponding results on robotic manipulation.}
\label{tab:setting_robotics}
\vspace{1em}
\centering
\begin{small}
\begin{tabular}{llllll}
\toprule
 & Model Size & Scales & \#Params & \#FLOPs & Success Rate \\
\midrule
\multirow{3}{*}{MVP} & Base & ($224^2$) & 86M & 17.5G & 43.8 \\
 & Base & ($224^2$, $448^2$) & 86M & 87.9G & 62.5 \\
 & Large & ($224^2$) & 307M & 61.6G & 50.0 \\
 \bottomrule
\end{tabular}
\end{small}
\vspace{1em}
\end{table}

\section{Derivation of Mutual Information}
\label{sec:appendix_derivation}

Denote the features from two models by $\vx \in \mathbb{R}^{d_x}$ and $\vy \in \mathbb{R}^{d_y}$ which follow the distribution $p(\rvx)$ and $p(\rvy)$, respectively. We make the simplest assumption that both the distribution and the conditional distribution of the features are isotropic gaussian distributions, \ie, $p(\rvy) \sim \mathcal{N}(\hat{\vmu}, \sigma^2 \mI)$ and $p(\rvy | \rvx) \sim \mathcal{N}(\hat{f}(\rvx), \sigma^{\prime2} \mI)$, where $f(\cdot)$ is a linear transform. The differential entropy and conditional differential entropy of $\rvy$ is $h(\rvy) = d_y \log \sigma + C$ and $h(\rvy | \rvx) = d_y \log \sigma^\prime + C$, where $C$ is a constant. The mutual information between features of two models is $I(\rvx; \rvy) = h(\rvy) - h(\rvy | \rvx) = d_y \log \sigma - d_y \log \sigma^\prime$.

When reconstructing the features $\rvy$ from another model's features $\rvx$, the optimal MSE loss would be $l = \min_{f} \frac{1}{d_y} E||\rvy - f(\rvx)||_2^2 = \frac{1}{d_y} E||\rvy - \hat{f}(\rvx)||_2^2 = \sigma^{\prime2}$. The optimal MSE loss of reconstructing $\rvy$ from a dummy constant vector would be $l_0 = \min_\vmu \frac{1}{d_y} E||\rvy - \vmu||_2^2 = \frac{1}{d_y} E||\rvy - \hat{\vmu}||_2^2 = \sigma^2$. Then we get the mutual information between $\rvx$ and $\rvy$ is $I(\rvx; \rvy) = d_y \log \sigma - d_y \log \sigma^\prime = - \frac{d_y}{2} \log \frac{\sigma^{\prime2}}{\sigma^2} \propto -\log \frac{l}{l_0}$.

\section{Results on ConvNeXt}
\label{sec:appendix_convnext}

\begin{figure}[ht]
  \includegraphics[width=1\linewidth]{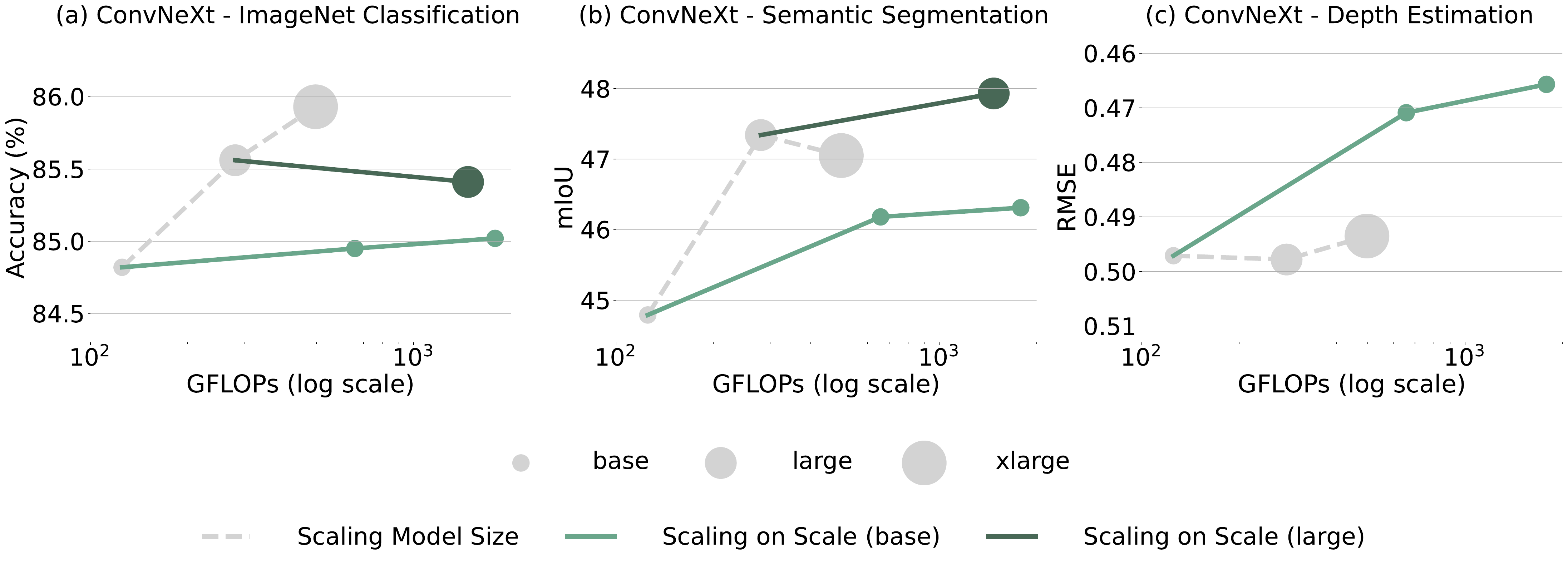}
  \vspace{-0.5em}
  \caption{\textbf{Comparison of \stwo scaling and model size scaling on ConvNeXt.} We evaluate three tasks: ImageNet classification, semantic segmentation, and depth estimation. For \stwo scaling (plotted in green curve), we test three sets of scales from single-scale (1x) to multi-scale (up to 3x), and we adjust each set of scale so that it matches the GFLOPs of the respective model size. Note that for specific models and tasks, we test \stwo scaling on both base and large models (plotted in light green and dark green curves separately).  }
  \label{fig:convnext_curve}
\end{figure}

To see if convolutional networks have similar behaviors as transformer-based models, we test ConvNeXt~\cite{liu2022convnet} models (per-trained on ImageNet-21k\footnote{\url{https://dl.fbaipublicfiles.com/convnext/convnext_base_22k_224.pth}}\textsuperscript{,}\footnote{\url{https://dl.fbaipublicfiles.com/convnext/convnext_large_22k_224.pth}}\textsuperscript{,}\footnote{\url{https://dl.fbaipublicfiles.com/convnext/convnext_xlarge_22k_224.pth}}) on three tasks: image classification, semantic segmentation, and depth estimation. We use ImageNet~\cite{russakovsky2015imagenet}, ADE20k~\cite{zhou2017scene}, and NYUv2~\cite{silberman2012indoor} datasets for each task. Similarly, we freeze the backbone and only train the task-specific head for all experiments, using a single linear layer, UPerNet~\cite{xiao2018unified}, and VPD depth decoder~\cite{zhao2023unleashing} as the decoder heads for three tasks, respectively. For model size scaling, we test the base, large, and xlarge size performance of ConvNeXt~\cite{liu2022convnet} model on each task. For \stwo scaling, we test three sets of scales including (1x), (0.5x, 1x, 2x), and (0.5x, 1x, 2x, 3x). 

The detailed curves are shown in Figure \ref{fig:convnext_curve}. We can see that in the depth estimation task (case (c)), \stwo scaling from base model significantly outperforms xlarge model with similar GFLOPs and only $0.25\times$ parameters. In the semantic segmentation task (case (b)), \stwo scaling from base model has less competitive result than larger models, while \stwo scaling from the large model outperforms the xlarge model with more GFLOPs but a smaller number of parameters. The ImageNet classification task (case (a)) is a failure case where \stwo scaling from both base and large model fail to compete with the xlarge model. From the observation above, we see that the convolutional networks show similar properties as transformer-based models: \stwo scaling has more advantages than model size scaling on dense prediction tasks such as segmentation and depth estimation while \stwo scaling is sometimes worse in image classification. This is possibly due to the fact that base and large model are not pre-trained with \stwo (see Section~\ref{sec:analysis}).

\section{Ablations of Model Design}
\label{sec:appendix_ablation}

We conduct the ablations on several designs of \stwowrapper. Specifically, (i) we first compare running vision model on sub-images split from the large-scale image with running on the large-scale image directly, and then (ii) we compare concatenating feature maps from different scales with directly adding them together.

Results for (i) are shown in Table \ref{tab:ablation_split}. We evaluate \stwowrapper with or without image splitting on ADE20k semantic segmentation. We test base and large baselines, as well as multi-scale base model with (1x, 2x) and (1x, 2x, 3x) scales separately. We can see that for (1x, 2x) scales, image splitting has better results than no splitting, which is due to image splitting makes sure the input to the model has the same size as in pre-training, and avoids performance degradation caused by positional embedding interpolation when directly running on large images. However, note that even running directly on large images, multi-scale base model still has better results than base and large models, which indicates the effectiveness of \stwo scaling. Furthermore, image splitting enjoys higher computational efficiency because it avoids the quadratic complexity of self-attention. Notice that without image splitting, the training will run into OOM error when using (1x, 2x, 3x) scales.

\begin{table}[h]
    \centering
    \caption{\textbf{Ablation of splitting large-scale images.} We compare splitting the large-scale image into regular-sized sub-images \vs running the model directly on the large image. We evaluate on ADE20k semantic segmentation. We can see that \stwo scaling with image splitting consistently outperforms directly running on the large image while being more compute-efficient. }
    \label{tab:ablation_split}
    \vspace{1em}
    \begin{small}
    \begin{tabular}{llcc}
        \toprule
            Model & Scales & Splitting & mIoU \\
        \midrule
            Base & $518^2$ &  & 54.8 \\
            Large & $518^2$ &  & 55.1 \\
            Base-\stwo & $518^2$, $1036^2$ & \xmark & 55.7 \\
            \rowcolor{lightgreen!40} Base-\stwo & $518^2$, $1036^2$ & \cmark & 56.3 \\
            Base-\stwo & $518^2$, $1036^2$, $1554^2$ & \xmark & OOM \\
            \rowcolor{lightgreen!40} Base-\stwo & $518^2$, $1036^2$, $1554^2$ & \cmark & 56.9 \\
        \bottomrule
    \end{tabular}
    \end{small}
    \vspace{0.5em}
\end{table}

Results for (ii) are shown in Table \ref{tab:ablation_concat}. We compare \stwowrapper with concatenating features from different scales with directly adding the features. We evaluate on ADE20k semantic segmentation with DINOv2 and OpenCLIP. On both models, concatenating, as done by default in \stwowrapper, has consistently better performance than adding the features.

\begin{table}[h]
    \centering
    \caption{\textbf{Ablation of how to merge features from different scales.} We compare concatenating features with adding features from different scales. Concatenating has consistently better performance.}
    \label{tab:ablation_concat}
    \vspace{1em}
    \begin{small}
    \begin{tabular}{llcc}
        \toprule
            Model & Scales & Merging & mIoU \\
        \midrule
            DINOv2-Base-\stwo & $518^2$, $1036^2$, $1536^2$ & add & 55.7 \\
            \rowcolor{lightgreen!40} DINOv2-Base-\stwo & $518^2$, $1036^2$, $1536^2$ & concat & 56.9 \\
            OpenCLIP-Base-\stwo & $256^2$, $512^2$, $1024^2$ & add & 51.4 \\
            \rowcolor{lightgreen!40} OpenCLIP-Base-\stwo & $256^2$, $512^2$, $1024^2$ & concat & 52.5 \\
        \bottomrule
    \end{tabular}
    \end{small}
    \vspace{0.5em}
\end{table}

\section{Throughput of Models with \stwo}

Previously we use FLOPs to measure the computational cost of different models. Since FLOPs is only a surrogate metric for the actual throughput of the models, here we compare the throughput of different models and verify if it aligns with FLOPs. Table \ref{tab:throughput} shows the results. We report the FLOPs and throughput of DINOv2 model with base, large, and giant size, as well as base size with scales of $(1\times)$, $(1\times, 2\times)$, and $(1\times, 2\times, 3\times)$. We test on base scales of $224^2$ and $518^2$. We can see that in general, the throughput follows the similar trends as FLOPs. For example, the base model with scales of ($224^2$, $448^2$, $672^2$) has the similar throughput as the giant model with scale of ($224^2$). The base model with scales of ($224^2$, $448^2$) has the about $0.8\times$ throughput as the large model with scale of ($224^2$). On base scale of $518^2$, the multi-scale base models with scales of $(1\times, 2\times)$, and $(1\times, 2\times, 3\times)$  have about $0.7\times$ throughput as the large and giant models, respectively.

\begin{table}[ht]
\caption{Comparison of FLOPs and Throughput.}
\label{tab:throughput}
\vspace{1em}
\centering
\begin{small}
\begin{tabular}{llll}
\toprule
Model Size & Scales & \#FLOPs & \begin{tabular}[c]{@{}l@{}}Throughput \\ (image/s)\end{tabular} \\
\midrule
Base & ($224^2$) & 17.6G & 138.5 \\
Base & ($224^2$, $448^2$) & 88.1G & 39.5 \\
Base & ($224^2$, $448^2$, $672^2$) & 246.0G & 16.5 \\
Large & ($224^2$) & 61.6G & 54.5 \\
Giant & ($224^2$) & 204.9G & 17.2 \\
\midrule
Base & ($518^2$) & 134.4G & 34.9 \\
Base & ($518^2$, $1036^2$) & 671.8G & 7.7 \\
Base & ($518^2$, $1036^2$, $1554^2$) & 1881G & 2.7 \\
Large & ($518^2$) & 460.9G & 11.8 \\
Giant & ($518^2$) & 1553G & 3.8 \\
 \bottomrule
\end{tabular}
\end{small}
\end{table}

\section{Additional Qualitative Results on V$^\ast$}
\label{sec:appendix_qualitative}

We show more qualitative results on the V$^\ast$ benchmark. We compare the performances of LLaVA-1.5 with \stwo scaling, original LLaVA-1.5~\cite{liu2023improved}, and GPT-4V~\cite{achiam2023gpt} on several examples in visual detail understanding (V$^\ast$~\cite{vstar}). Similarly, for LLaVa-1.5 with \stwo scaling, we use Vicuna-7B~\cite{chiang2023vicuna} as LLM and OpenAI CLIP as the vision backbone and apply \stwo scaling on the vision backbone.

\begin{figure*}[h!]
    \centering
    \begin{subfigure}[t]{0.45\textwidth} %
        \centering
        \includegraphics[width=\linewidth]{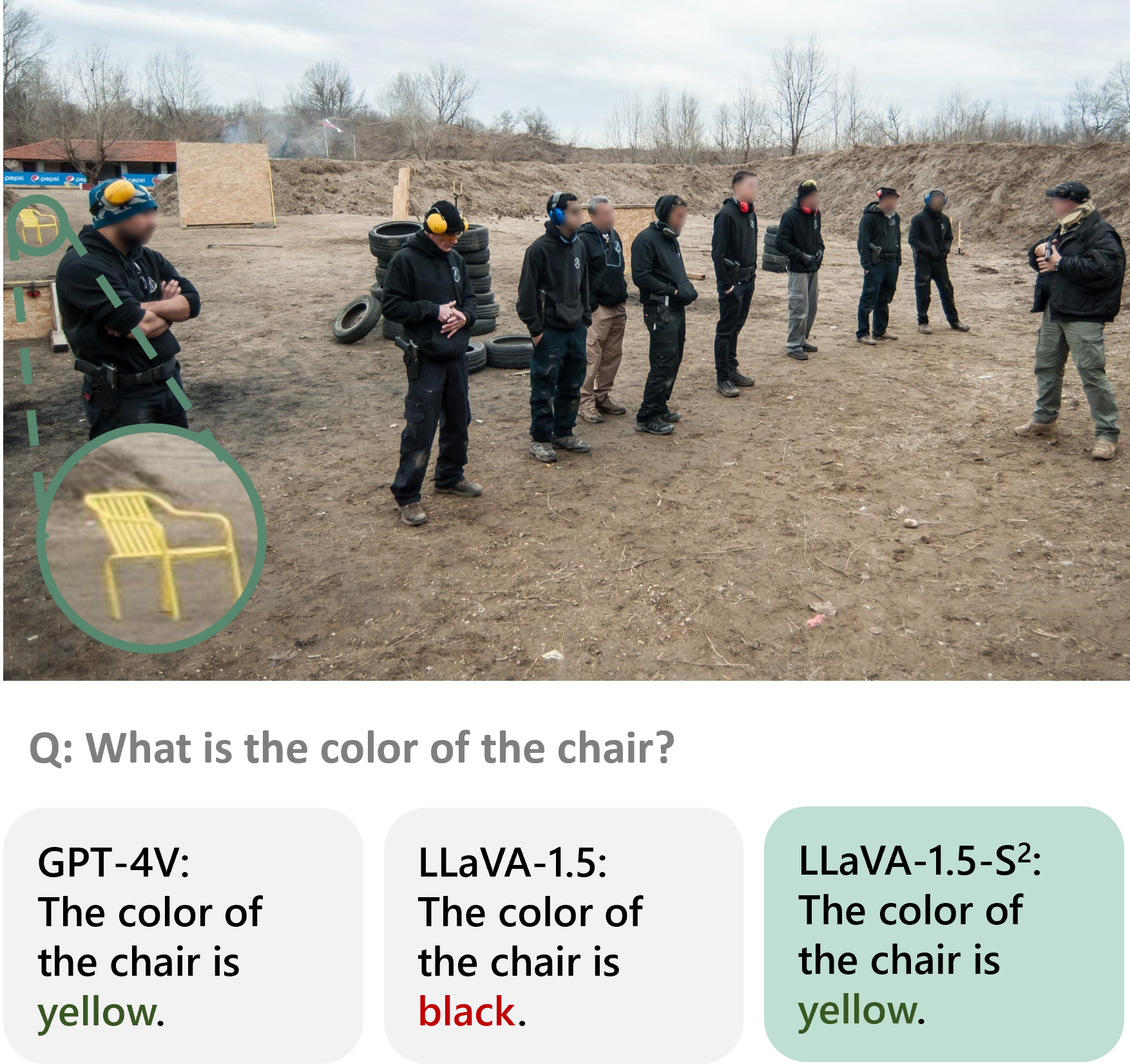} %
        \caption{What is the color of the chair?} %
    \end{subfigure}
    ~ %
    \begin{subfigure}[t]{0.45\textwidth}
        \centering
        \includegraphics[width=\linewidth]{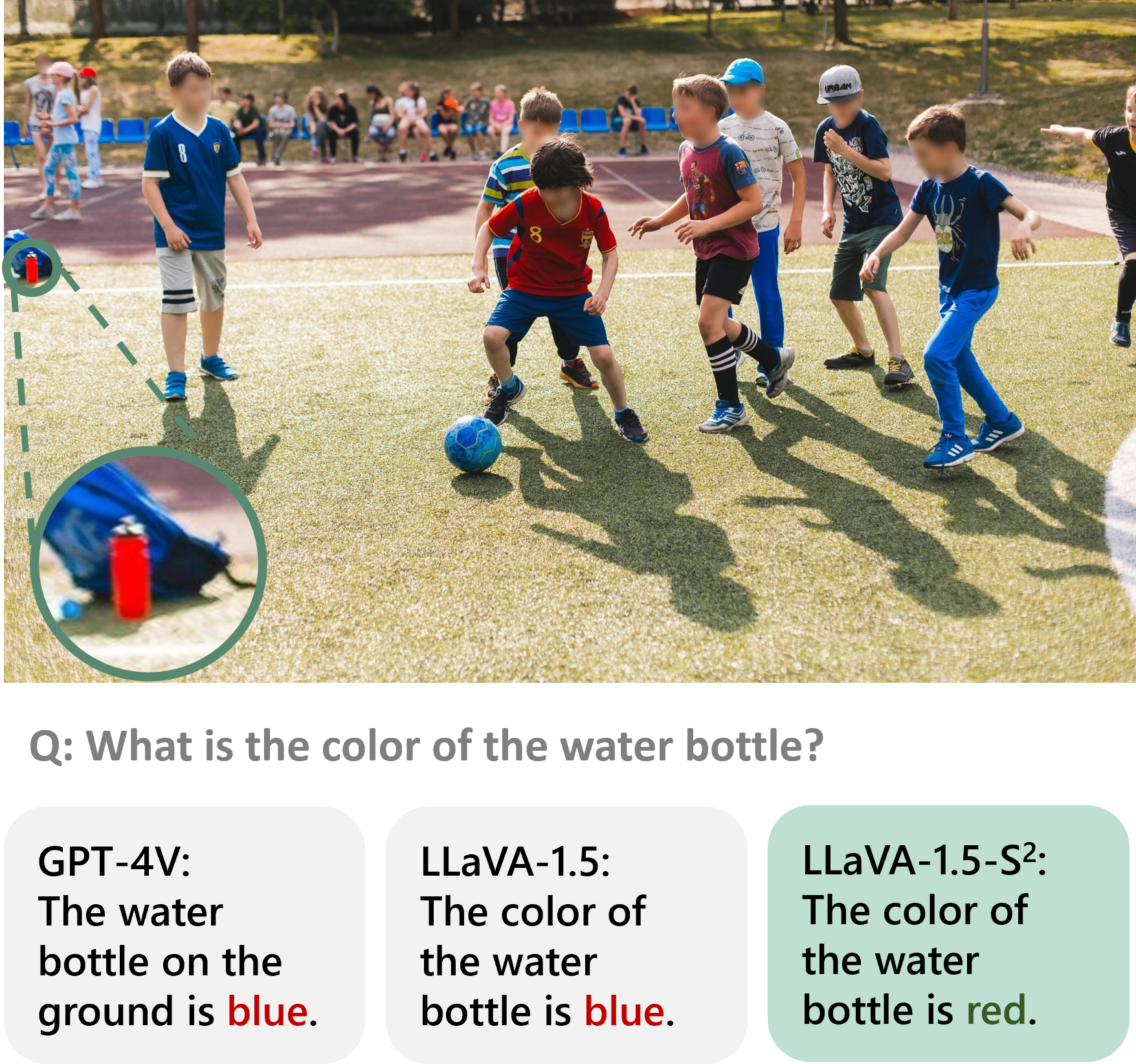}
        \caption{What is the color of the water bottle?}
    \end{subfigure}
    \par\medskip %
    \begin{subfigure}[t]{0.45\textwidth}
        \centering
        \includegraphics[width=\linewidth]{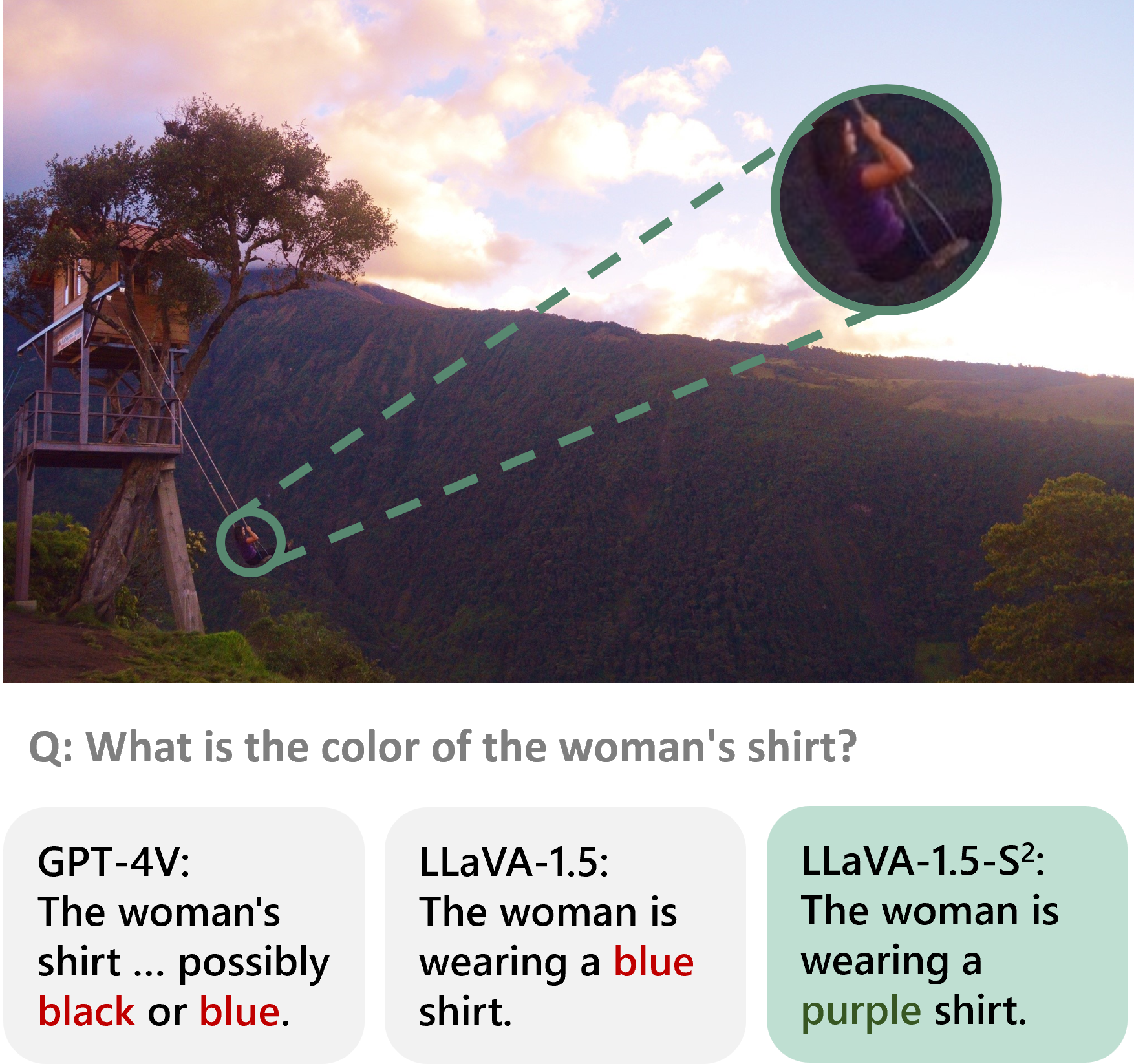}
        \caption{What is the color of the woman's shirt?}
    \end{subfigure}
    ~
    \begin{subfigure}[t]{0.45\textwidth}
        \centering
        \includegraphics[width=\linewidth]{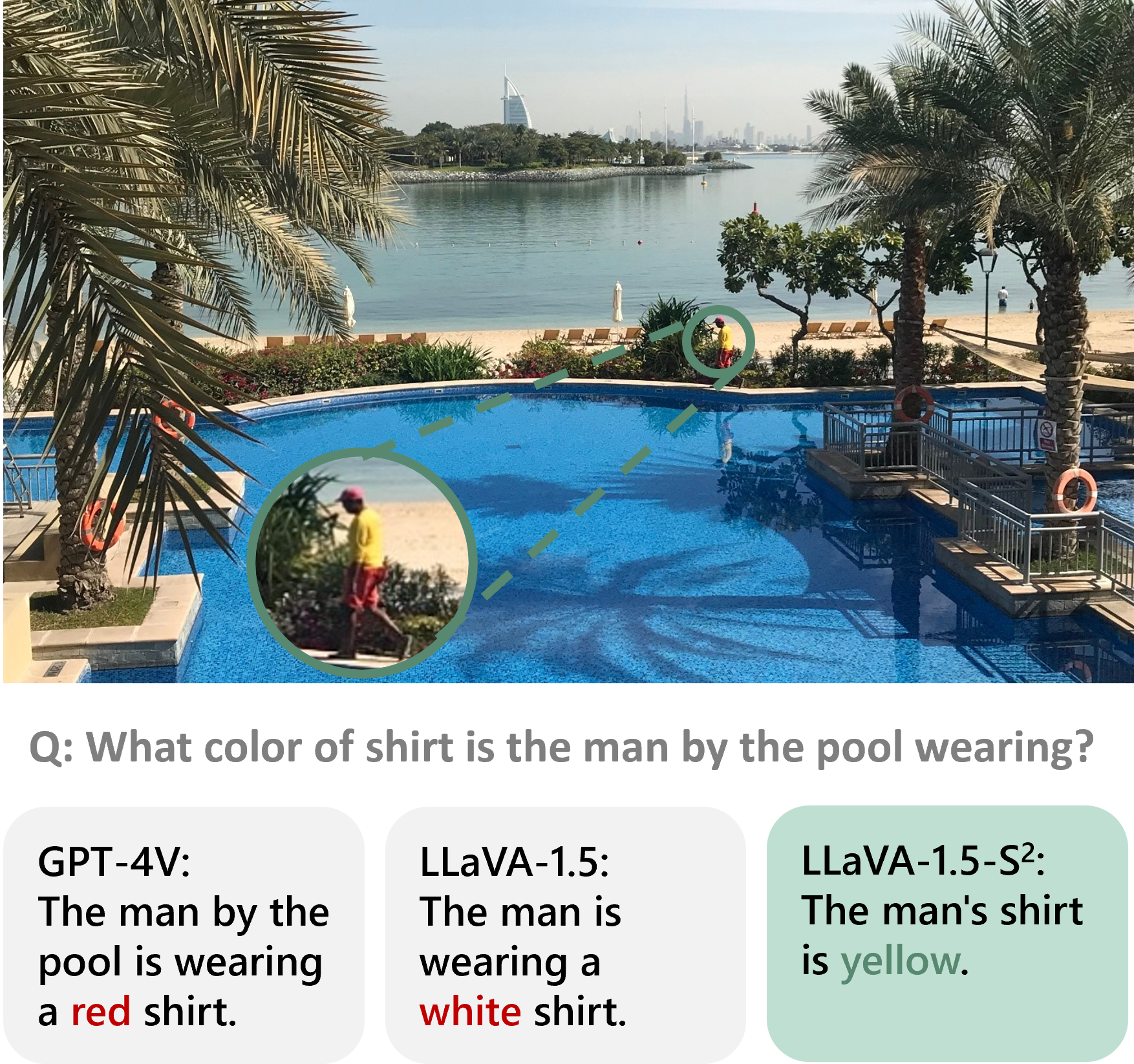}
        \caption{What color of shirt is the man by the pool wearing?}
    \end{subfigure}
    \par\medskip %
    \begin{subfigure}[t]{0.45\textwidth}
        \centering
        \includegraphics[width=\linewidth]{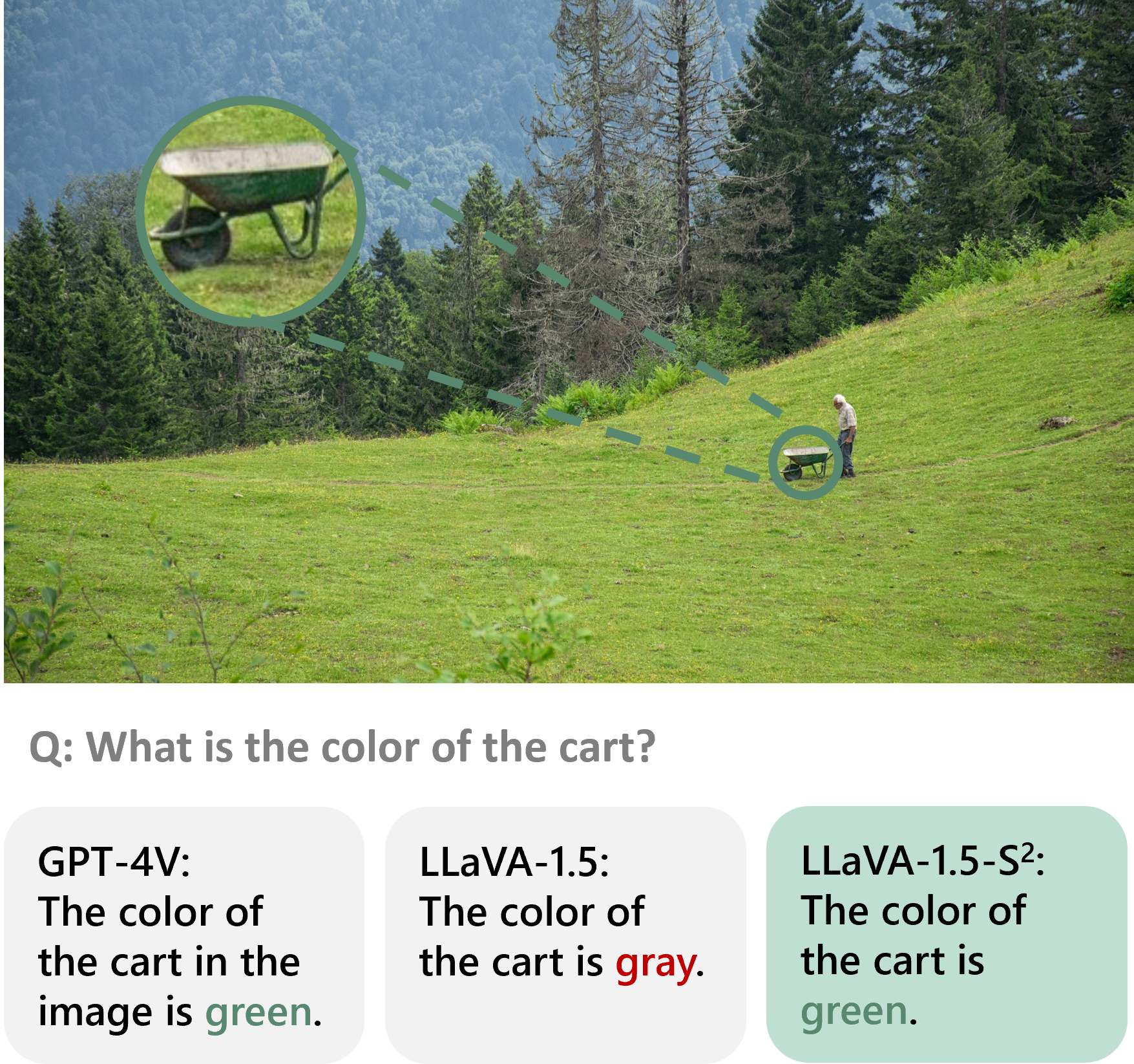}
        \caption{What is the color of the cart?}
    \end{subfigure}
    ~
    \begin{subfigure}[t]{0.45\textwidth}
        \centering
        \includegraphics[width=\linewidth]{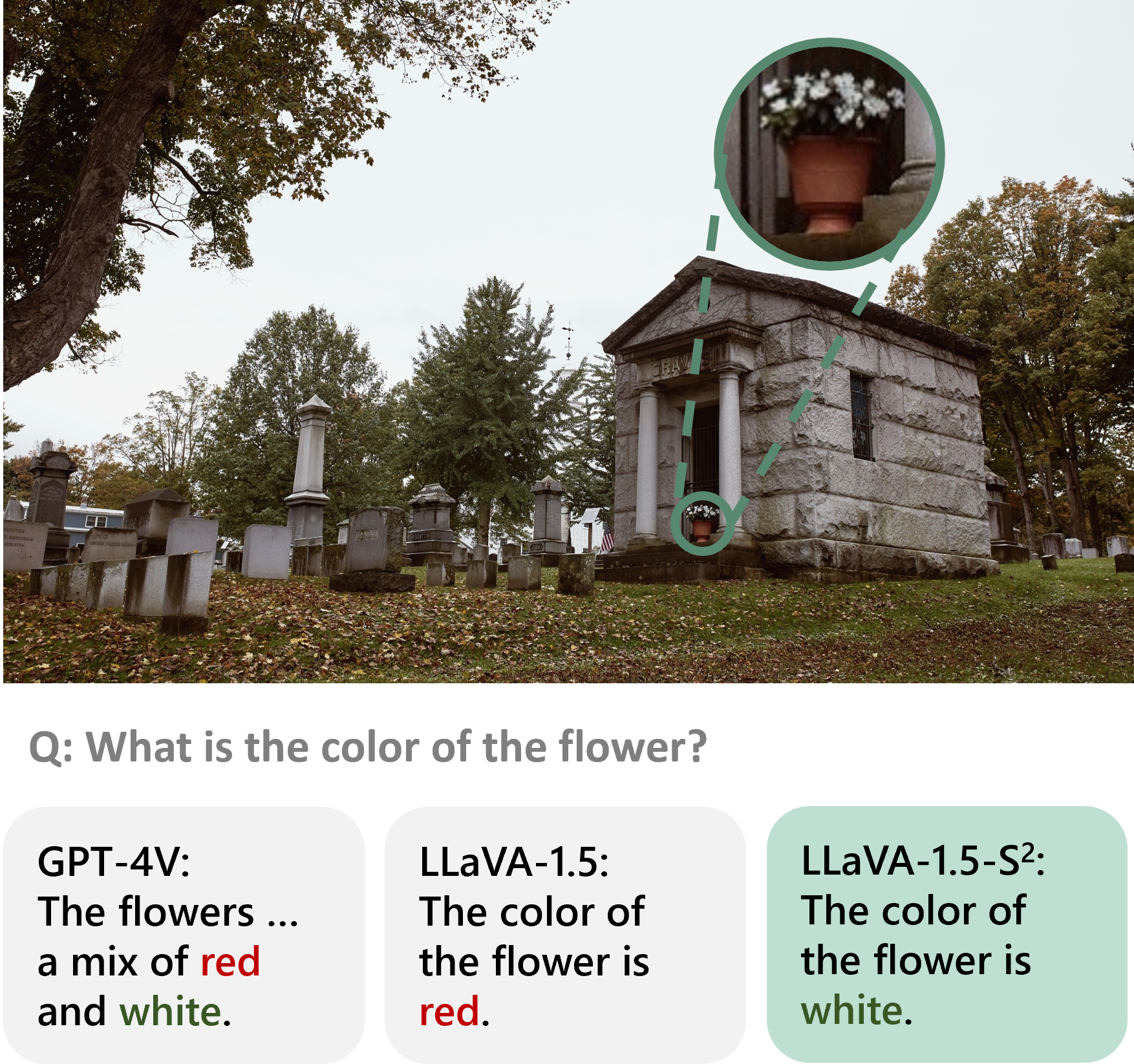}
        \caption{What is the color of the flower?}
    \end{subfigure}
    
    \caption{\textbf{Examples of LLaVA-1.5 with \stwo scaling on the V$^\ast$ benchmark,} demonstrating its extreme ability in recognizing fine-grained details of an image.} %
    \label{fig:vstar}
\end{figure*}

In Figure \ref{fig:vstar}, we see various examples that demonstrate the capabilities of different MLLMs. For instance, in example (f), the query is about the color of the flowers, which only occupy around 670 pixels in the $2550\times1500$ image. Here, LLaVA-1.5-\stwo correctly identifies the color as 'white'. However, LLaVa-1.5 fails to capture the correct color and recognizes it as 'red', which is actually the color of the flowerpot. On the other hand, GPT-4V recognizes the color as 'a mix of red and white', indicating that it cannot distinguish the subtle differences between the flowerpot and flowers.

In another example (c), the query is about the color of the woman's shirt. Here, the size of the woman's figure is small, and the purple color of the shirt is very similar to the dark background color. In this case, LLaVA-1.5-\stwo correctly identifies the color of the shirt as 'purple', while both LLaVA-1.5 and GPT-4V mistakenly identify the color of the shirt as 'black' or 'blue', which is the color of the background.

The above examples highlight the difference in performance between LLaVA-1.5-\stwo, LLaVA-1.5 and GPT-4V. LLaVA-1.5-\stwo distinguishes itself through its heightened sensitivity and enhanced precision in visual detail understanding. This advanced level of detail recognition can be attributed to the \stwo scaling applied to its vision backbone, which significantly augments its ability to analyze and interpret subtle visual cues within complex images.

\newpage

\end{document}